%% file: acl_latex.tex
\pdfoutput=1
\documentclass[11pt]{article}

\usepackage[final]{acl}

\usepackage{times}
\usepackage{latexsym}

\usepackage[T1]{fontenc}
\usepackage[utf8]{inputenc}

\usepackage{microtype}

\usepackage{inconsolata}

\usepackage{graphicx}

\usepackage{amssymb}  % for null document
\usepackage{float}  % At EACL 2024
\usepackage[most]{tcolorbox} % Load the tcolorbox package with additional libraries
\usepackage{xcolor} % Required for specifying colors
\usepackage{amsmath}
\definecolor{darkgray}{gray}{0.3}
\definecolor{lightgray}{gray}{0.9}
\newtcolorbox{mybox}[1][]{
  colback=darkgray,
  colframe=darkgray,
  coltext=white,
  boxsep=5pt,
  arc=0pt,
  outer arc=0pt,
  #1
}
\usepackage{booktabs} % For prettier tables
\usepackage{array}    % For table wrapping and fixed-width columns

\usepackage{multirow}

\usepackage{kotex}
\usepackage{subcaption}
\usepackage{booktabs}
\usepackage{multirow}
\usepackage{makecell}
\usepackage{graphics}
\usepackage{bbm}
\usepackage{lipsum}% http://ctan.org/pkg/lipsum
\usepackage[english]{babel}
\usepackage{algorithm}
\usepackage[noend]{algpseudocode}
\usepackage{enumitem}
\usepackage{bm}
\makeatletter
\algnewcommand{\LineComment}[1]{\Statex \hskip\ALG@thistlm \(\triangleright\) #1}
\makeatother
\title{Correcting Negative Bias in Large Language Models through \\ Negative Attention Score Alignment}

\author{Sangwon Yu$^{1}$\thanks{Equal Contribution} \hspace{2mm}  Jongyoon Song$^{1}$\footnotemark[1] \hspace{2mm}    Bongkyu Hwang$^{2}$   \hspace{2mm}   Hoyoung Kang$^{2}$ \hspace{2mm}  Sooah Cho$^{2}$ \\ {\bf Junhwa Choi$^{2}$} \hspace{2mm}  {\bf Seongho Joe$^{2}$} \hspace{2mm}  {\bf Taehee Lee$^{2}$}\hspace{2mm}  {\bf Youngjune Gwon$^{2}$}\hspace{2mm}  {\bf Sungroh Yoon$^{1,3}$\Thanks{\hspace{0.1em} Corresponding author}} \\
   $^{1}$Department of Electrical and Computer Engineering, Seoul National University \\
   $^{2}$Samsung SDS, Korea \\
   $^{3}$AIIS, ASRI, INMC, ISRC, and IPAI, Seoul National University \\
   {\tt\footnotesize \{dbtkddnjs96, coms1580, sryoon\}@snu.ac.kr} \\
   {\tt\footnotesize \{bongkyu.hwang, hoyoung.kang, sooah.cho, jh496.choi, drizzle.cho, taehee23.lee, gyj.gwon\}@samsung.com}}

\begin{document}
\maketitle
\begin{abstract}
A binary decision task, like yes-no questions or answer verification, reflects a significant real-world scenario such as where users look for confirmation about the correctness of their decisions on specific issues. In this work, we observe that language models exhibit a negative bias in the binary decisions of complex reasoning tasks. Based on our observations and the rationale about attention-based model dynamics, we propose a negative attention score (NAS) to systematically and quantitatively formulate negative bias. Based on NAS, we identify attention heads that attend to negative tokens provided in the instructions as answer candidate of binary decisions, regardless of the question in the prompt, and validate their association with the negative bias. Additionally, we propose the negative attention score alignment (NASA) method, which is a parameter-efficient fine-tuning technique to address the extracted negatively biased attention heads. Experimental results from various domains of reasoning tasks and large model search space demonstrate that NASA significantly reduces the gap between precision and recall caused by negative bias while preserving their generalization abilities.
\end{abstract}

\input{sections/1_Introduction}

\input{sections/2_Related_Work}

\input{sections/3_Negative_Bias_in_LLM}
\input{sections/4_Method}

\input{sections/5_Experiment}

\input{sections/6_Analysis}
\input{sections/7_Discussion}

\input{sections/8_Conclusion}

\input{sections/Limitations}

\section*{Acknowledgements}
This work was supported by the National Research Foundation of Korea (NRF) grant funded by the Korea government (MSIT) (No. 2022R1A3B1077720, No. 2022R1A5A708390811), 
Institute of Information \& communications Technology Planning \& Evaluation (IITP) grant funded by the Korea government(MSIT) [NO.RS-2021-II211343, Artificial Intelligence Graduate School Program (Seoul National University)],
the BK21 FOUR program of the Education and Research Program for Future ICT Pioneers, Seoul National University,
Samsung SDS Co., Ltd, 
and a grant from the Yang Young Foundation.

\bibliography{anthology,custom}
% \bibliography{custom}

\clearpage
\appendix
\input{sections/A_data_Statistics}
\input{sections/A_statistical_analysis}

\input{sections/C_Instructions}
\input{sections/details_of_binary_decision_data_construction}

\input{sections/details_of_parametric_sample_selection}

\input{sections/NASA_Algorithm}
\input{sections/neg_response_analysis}
\input{sections/response_shift_ratio_analysis}
\input{sections/6.2_Generalization_to_Universal_Binary_Decision}
\input{sections/6.3_Transferability_across_Various_Instructions}
\input{sections/Ablation_Study}

\input{sections/confidence_histogram_analysis}
\input{sections/D_Samples}
\input{sections/Computation}
\input{sections/License}
\input{sections/E_AI_Assistant}

\end{document}

%% file: sections/1_Introduction.tex
\section{Introduction} \label{1.introduction}

Recent advancements in large language models (LLMs) have enabled complex reasoning tasks executed by understanding user instructions (\citealp{ouyang2022training}; \citealp{touvron2023llama}; \citealp{achiam2023gpt}; \citealp{jiang2023mistral}). 
% LLMs have demonstrated an exceptional ability to execute a variety of tasks by understanding user instructions, especially those requiring complex reasoning.
As the capabilities of LLMs have expanded rapidly, research has intensified to analyze their characteristics and inherent issues. 
One of the major issues is the generation of factually incorrect content, known as ``hallucination'', which significantly degrades the reliability of LLM-based services (\citealp{zhang2023siren}; \citealp{xu2024hallucination}; \citealp{huang2023survey}).

The hallucination problem in LLMs can be attributed to factors such as parametric knowledge, overconfidence, and biases (\citealp{zhang2023siren}). 
Studies have been conducted to understand the decision-making mechanisms in LLMs from a parameter perspective. 
For instance, the ``logit lens'' technique allows for the interpretation of the model’s reasoning process along the layers, exploiting the hidden representations from intermediate layers (\citealp{belrose2023eliciting}; \citealp{ferrando-etal-2023-explaining}). 
This technique has been utilized to analyze model characteristics in various scenarios such as in-context learning. 
On the other hand, \citet{yuan2024whispers} focus on analyzing hallucination phenomenon based on attention heads.
However, the causes and mechanisms of hallucination in LLMs vary by type of task, indicating that extensive further study is still necessary.

In this paper, we aim to identify biases that emerge when LLMs respond to questions requiring a binary decision, such as affirmation or negation. 
We observe a negative bias in LLMs in yes-no question-answering (QA) and answer verification tasks that demand complex reasoning, such as mathematical or logical reasoning.
% Specifically, 우리는 이러한 binary decision tasks에서 LLMs가 전반적으로 high precision and low recall performance를 보임을 관측한다. 또한 우리는 negative decision에 대한 prediction confidence가 positive decision의 경우보다 significantly high함을 확인한다. 이는 LLMs가 지나치게 신중하게 positive decision을 수행함을 의미하며, 다른 말로는 negative decision을 남발하는 shortcut에 해당하는 현상이 발생하고 있다고 할 수 있다.
% This negative bias, which implies a discrepancy between the model's reasoning ability and accuracy, potentially decreases the reliability of model predictions.
Specifically, we observe that LLMs generally exhibit high precision but low recall in binary decision tasks. Additionally, we find that the prediction confidence for negative decisions is significantly higher than for positive decisions. This suggests that \textit{LLMs tend to be overly cautious when making positive decisions, leading to a phenomenon where the model excessively favors negative decisions as a form of shortcut}. This negative bias, which indicates a discrepancy between the model's reasoning ability and accuracy, may ultimately reduce the reliability of its predictions.

From the rationale that the model allocates higher attention to the negative candidates among the binary answer options presented in the user prompt, we propose a \textbf{n}egative \textbf{a}ttention \textbf{s}core (\textbf{NAS}) to systematically probe these negatively biased attention heads in LLMs.
Based on NAS, we detect attention heads that attend to tokens associated with negation in an instruction.
% independent of the user query when the model is required to generate a positive response. 
Our findings also confirm the existence of attention heads that predominantly contribute to manifesting this negative bias, regardless of the given query.
To address the identified attention heads, we introduce a parameter-efficient fine-tuning technique named \textbf{NAS} \textbf{a}lignment (\textbf{NASA}).

In the NASA framework, we first construct a probing set in the form of a binary decision task to extract negative attention heads. This probing set is derived from an existing short-answer QA dataset. Next, we apply an incremental fine-tuning process to the negative attention heads identified based on NAS. During this process, each head undergoes fine-tuning using the probing set, while NASA includes periodic monitoring of NAS, which automatically schedules early stopping and update cancellations. This pipeline ensures that tuning is conducted only up to an appropriate extent and in the correct sequence, effectively mitigating negative bias without compromising the model’s existing capabilities.
% In NASA framework, 우리는 우선 the negative attentions heads를 extract하기 위한 binary decision task 형태 probing set을 existing short-answer QA dataset으로부터 construct 한다. 이후 probing set에 대한 NAS를 기반으로 추출된 negative attention heads들에 대해 각 head의 NAS를 기반으로 한 incremental fine-tuning 과정을 적용한다. Head 하나씩에 대해서 probing set을 통한 fine-tuning을 진행되는 과정에서 NASA는 includes periodic monitoring of NAS, which automatically schedules early stopping and update cancellations. 이러한 pipeline은 적절한 순서로 적절한 지점까지만 tuning을 진행하여 기존 capability의 손상 없이 negative bias만을 효과적으로 완화시키는 데에 기여한다.
% the model is incrementally fine-tuned, starting with the most biased attention head. 
% The method includes periodic monitoring of NAS, which automatically schedules early stopping and update cancellations.

For the evaluation of NASA, we measure the performance of four LLMs: LLaMA3-8B-Instruct (\citealp{touvron2023llama}), Mistral-7B-Instruct-v0.3 (\citealp{jiang2023mistral}), Gemma-1.1-7b-it (\citealp{team2024gemma}), and Qwen2-7B-Instruct (\citealp{yang2024qwen2}) as the search space. We conduct a comprehensive evaluation across multiple reasoning tasks, including multi-hop QA: StrategyQA (\citealp{geva2021strategyqa}), MuSiQue (\citealp{trivedi2022musique}), mathematical reasoning: GSM8k-Rephrased and MATH-Rephrased subset in MetaMATH (\citealp{yu2023metamath}), and logical reasoning: AR-LSAT (\citealp{zhong2021arlsat}). 

 Experimental results demonstrate that our method NASA significantly reduces the gap between precision and recall while maintaining or even improving accuracy and F1 score. This indicates that the NASA method successfully mitigates negative bias without compromising overall performance. Additionally, we confirm that the NASA-tuned model effectively preserves general reasoning performance beyond binary decision-making and exhibits improvements in calibration. Furthermore, NASA demonstrates robust performance across various prompting formats, including few-shot settings.
% For evaluation of NASA, 우리는 four LLMs: LLaMA3-8B-Instruct (\citealp{touvron2023llama}), Mistral-7B-Instruct-v0.3 (\citealp{jiang2023mistral}), Gemma-1.1-7b-it (\citealp{team2024gemma}), Qwen2-7B-Instruct (\citealp{yang2024qwen2})을 search space로 하여 각 model의 성능을 측정한다. 우리는 multi-hop QA: StrategyQA (\citealp{geva2021strategyqa}), MuSiQue (\citealp{trivedi2022musique}), mathematical reasoning: MetaMATH (GSM8k-Rephrased and MATH-Rephrased; \citealp{yu2023metamath}), logical reasoning: AR-LSAT (\citealp{zhong2021arlsat}) task에 대해서 comprehensive한 평가를 진행한다.
% Experimental results demonstrate that our method significantly reduces the gap between precision and recall while maintaining or enhancing the accuracy and F1 score. 이는 NASA method가 성공적으로 negative bias만을 mitigate했음을 의미한다.
% 추가적으로, 우리는 NASA-tuned model이 binary decision 범위 밖의 general reasoning performance를 잘 유지하고 있으며, calibration 측면에서 향상되었음을 확인한다. 이 밖에도 NASA는 few-shot을 포함한 다양한 prompt 형태에 대해 robust한 성능을 보인다.
% negative bias while maintaining general reasoning performance. 
% Additionally, we observe an improvement in calibration for binary decision tasks in LLMs.
% \\

Our contributions can be summarized as follows:
\begin{itemize}
    \item We observe that LLMs exhibit a negative bias in binary decision tasks requiring complex reasoning and demonstrate its association with attention heads.
    \item We propose negative attention score (NAS), a metric and framework that allows for systematic probing of attention heads involved in negative bias regardless of the input query. 
    \item Based on NAS, we introduce NAS alignment (NASA), a parameter-efficient fine-tuning technique that effectively addresses negative bias by tuning the extracted negative attention heads in a proper order.
\end{itemize}

%% file: sections/2_Related_Work.tex
\section{Related Work} \label{2.related work}

\textbf{Hallucination problems in complex reasoning tasks}
% LLMs are demonstrating capabilities not only in typical natural language processing tasks but also in tasks that require complex thought processes. 
The emergent abilities of LLMs, such as the chain-of-thoughts, have significantly enhanced the performance of complex reasoning (\citealp{wei2022chain}; \citealp{kojima2022large}; \citealp{wang2022self}; \citealp{madaan2023self}). 
However, the increased complexity of tasks and pipelines has made the hallucination problem in complex reasoning tasks challenging to analyze and resolve, which remains an active area of research (\citealp{zhang2023siren}).
With the advent of benchmarks vulnerable to hallucinations, recent LLMs are laying the groundwork for analyzing the causes of LLM hallucinations based on substantial data which requires complex reasoning abilities (\citealp{geva2021strategyqa}; \citealp{trivedi2022musique};\citealp{cobbe2021training}). 
% For instance, StrategyQA () and MuSiQue () involve queries and contexts that demand multi-step reasoning from LLMs, while GSM8k () involves queries that require mathematical reasoning skills. 

% In this paper, we aim to understand hallucinations from the perspective of model parameters, using the currently employed complex reasoning datasets. 
% Specifically, we focus on binary decision tasks involving affirmation and negation to identify new characteristics of the model.

\textbf{Analysis on roles and knowledge in model parameters}
As the capacity and complexity of LLMs increase, the need to understand and interpret these models is becoming more crucial. 
In particular, there is active research into analyzing parametric knowledge and model behavior that influence the phenomenon of model hallucination.
\citet{meng2022locating} propose a framework to specify the location where the knowledge of the model is stored, while \citet{belrose2023eliciting} and \citet{yang2024large} research to interpret the inference process of the model. 
\citet{yuan2024whispers} suggest a method to detect model parameters involved in the false premise hallucination phenomenon of LLMs.

% Similarly, we demonstrate that the model's negative bias is related to the attention heads. 
% We explore the existence and characteristics of query-agnostic negative bias in attention heads.
% We further demonstrate that this can be significantly addressed through a parameter-efficient fine-tuning.

% There is active research on analyzing the internal module dynamics to enhance the interpretability of LLMs.

\input{figures/precision_recall_}
\input{figures/conf_hists_}

Recent studies have shown that attention heads in language models play a variety of roles such as knowledge recalling (\citealp{zheng2024attention}).
For example, \citet{jin-etal-2024-cutting} find the existence of attention heads in LLMs that either recall parametric knowledge or retrieve information from external contexts and \citet{ijcai2024p43} propose a method to identify attention heads that increase vulnerability to adversarial attacks.

In this paper, we identify the negative bias in LLMs that leads to hallucinations in binary decision tasks requiring complex reasoning.  
We focus on the attention heads in LLMs and propose a framework to detect and address the attention heads responsible for the negative bias.

% \textbf{Analysis on attention heads in large language models}
% LLM의 interpretability를 증진시키기 위해 내부 module의 dynamics를 분석하는 연구가 활발히 이루어지고 있다.

%% file: figures/precision_recall_.tex
\begin{figure*}[!t]
  \includegraphics[width=0.95\textwidth]{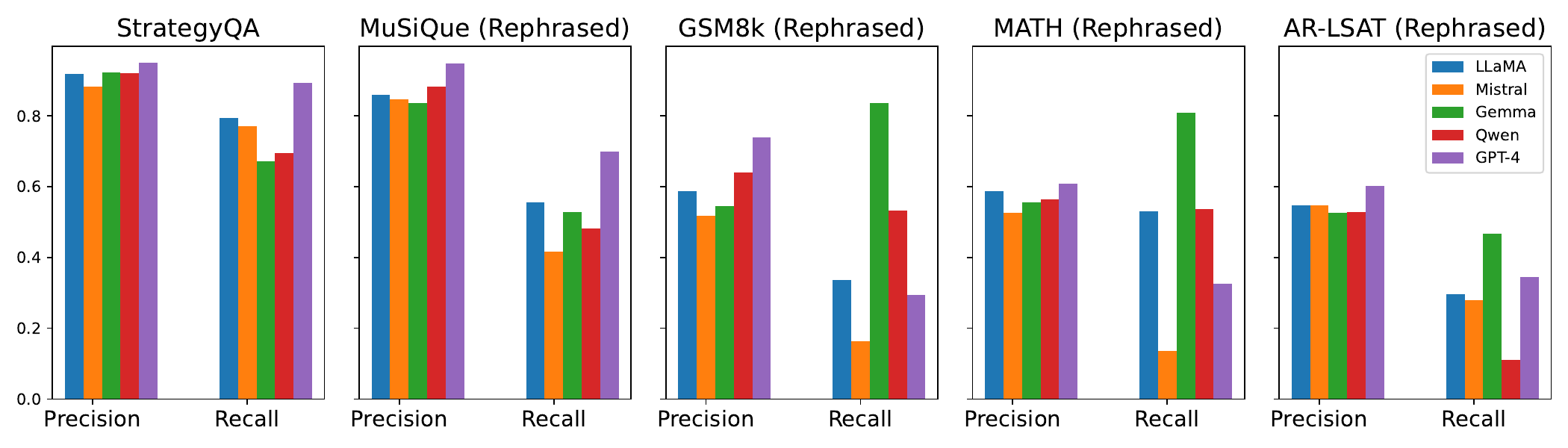}
  \caption{Precision and recall of LLMs in complex reasoning tasks.}
  \label{fig:precision_recall}
\end{figure*}

%% file: figures/conf_hists_.tex
\begin{figure*}[!t]
  \includegraphics[width=\textwidth]{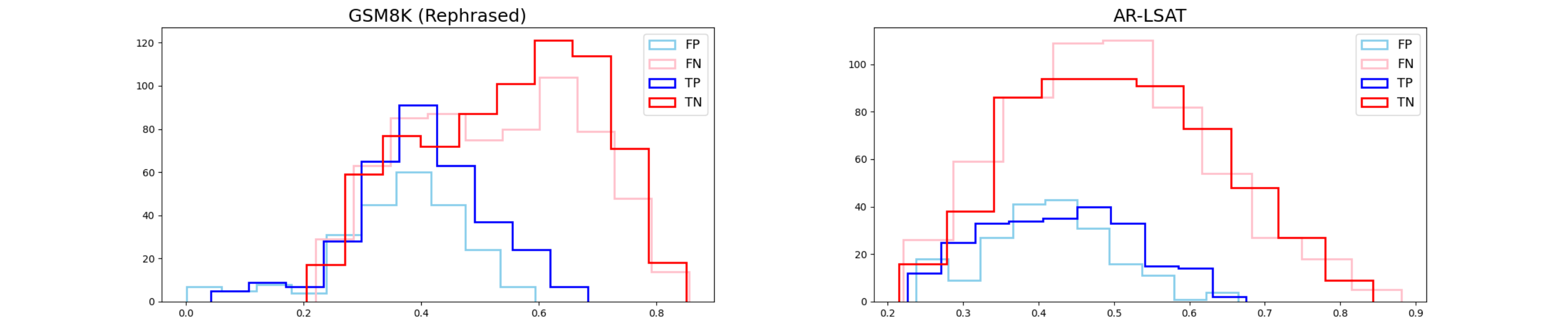}
  \caption{Confidence histograms of LLaMA3-8B-Instruct model for the rephrased GSM8K and AR-LSAT samples.}
  \label{fig:conf_hist}
\end{figure*}

%% file: sections/3_Negative_Bias_in_LLM.tex
\section{Negative Bias in LLMs} \label{3.problem}
 % Binary decision tasks, where the user requests for the model to generate a positive or negative response based on the input provided, play a significant role in the interaction between model agents and users. For instance, a user may inquire whether a statement formed through various contexts is true. Also, a user may seek validation of the correctness of their proposed solutions to problems requiring complex logical reasoning. Such fundamental interactions, represented as Yes-No decisions, act as the basic block of the high-dimensional reasoning tasks that need step-by-step sub-tasks.
% \input{figures/precision_recall_}
% \input{figures/conf_hists_}

% \input{figures/precision_recall_}
% \input{figures/conf_hists_}

A binary decision task, such as yes-no QA or answer verification, represents a major real-world scenario where users seek confirmation on whether their decisions regarding specific problems are correct. 
 For instance, a user may inquire whether a statement formed through various contexts is true.
 Also, a user may seek validation of the correctness of their proposed solutions to problems requiring complex logical reasoning.
 Such fundamental interactions, represented as yes-no decisions, act as the basic block of the high-dimensional reasoning tasks that need step-by-step sub-tasks.
% This acts as the basic block of the high-dimensional reasoning tasks that need step-by-step sub-tasks.
 Despite this significance of the binary decision task, many complex reasoning benchmark datasets are not structured in a binary decision format.
 In this section, we first transform existing complex reasoning tasks into yes-no binary decision tasks.
 % We then conduct statistical observations and quantitative analyses on how LLMs perform in these binary decision scenarios and 이들의 decision에서 meaningful negative bias를 관측한다.
 % After that, 우리는 LLMs의 negative bias를 formulate하기 위한 systemic framework, negative attention score를 제안한다.
 We then conduct statistical observations and quantitative analyses to examine how LLMs perform in these binary decision scenarios, observing a meaningful negative bias in their decisions. Following this, we propose a systematic framework, the \textbf{N}egative \textbf{A}ttention \textbf{S}core (\textbf{NAS}), to formulate and quantify the negative bias exhibited by LLMs.
\input{sections/3.1_Statistical_Observation}
\input{sections/3.2_Formulation_NAS}

%% file: sections/3.1_Statistical_Observation.tex
\subsection{Statistical Observation} \label{3.1.stat_obs}
% First, we experimentally demonstrate that negative biases occur in various LLMs during complex reasoning tasks.
% 우리는 multi-step reasoning 및 mathematical reasoning을 요구하는 yes-no question answering을 통해negative bias 현상을 관측한다.

% In this work, 우리는 complex reasoning task 환경에서 large language model의 binary decision에 대한 작동 기전을 explore한다. User가 제공하는 입력에 대해서 긍정 또는 부정을 나타내는 binary decision task는 LLM agent와 user 간 interaction에서 굉장히 큰 비중을 차지하고 있다. 가령 user는 다양한 문맥을 통하여 조합된 statement가 사실인지 질의할 수 있고, 또한 복잡한 논리 추론 과정이 필요한 문제에 대해 자신이 제시하는 solution의 correctness를 확인받고자 할 수 있다. 이와 같이 Yes-No로 대표되는 기본적인 상호작용은 고차원적인 reasoning task로 갈수록 더욱 근본적인 중요성을 담당하고 있다고 볼 수 있다. In this section, 우리는 기존에 존재하는 complex reasoning task를 Yes-No binary decision task로 transform한 뒤, 다양한 large language model들이 해당binary decision task에서 어떠한 양상을 보이는지에 대한 통계적 관측 및 정량적 분석을 진행한다.

\subsubsection{Setup for Binary Decision Task}\label{subsubsec:binary_decision}
% Many of complex reasoning benchmark set은 yes-no question answering 형태가 아니다.
% 우리는 GPT-4를 이용하여 general question answering dataset의 target answer를 Yes 또는 No로 modify하였다.
% 구체적으로, 우리는 question과 correct answer, 그리고 given context를 참조하여 wrong answer를 출력하도록 prompt했다.
% 우리는 context를 함께 제공했을 때, context 내에 있는 entity를 wrong answer로서 출력하는 성향을 확인했다, which 오답 여부를 판별하기 challenging하게 한다.
% 답이 yes인 prompt는 question의 정답이 ground truth가 맞는지 물어봄으로써, no인 prompt는 question의 정답이 contaminated answer인지를 물어봄으로써 이루어진다.
\paragraph{Transformation to binary decision task}
% In this work, 우리는 두 가지 방식을 사용하여 기존의 일반적인 QA reasoning task를 binary decision task: Yes-No QA 형태로 변환한다. 첫째로, 우리는 기존 query에 대한 label을 query 뒤에 되묻는 형태의 rule-based transformation을 수행한다. 예를 들어 "What is 1+1?"이라는 기존 query를 우리는 "What is 1+1? Is the answer 2?"의 모양으로 변형한다. 이러한 방식은 간편하다는 장점이 있지만 오직 정답이 ``Yes''인 case로만 변형할 수 있다는 한계가 존재한다. 정답이 ``No''인 경우로 변형하기 위해서는 기존 sample에서 제시된 query와 answer에 더해서 wrong answer가 필요하다. 이를 위해 우리는 existing large language model (LLM)에 prompting을 하여 적절한 wrong answer를 성공적으로 획득한다. 그 뒤에, 앞서 언급한 rule-based 방식을 통해 기존 QA sample을 정답이 No인 binary decision sample로 변형한다. 둘째로, 우리는 LLM에 대한 prompting을 통하여 기존 query를 자연스러운 binary decision query로 변환한다. Rule-based transformation보다 더 realistic하고 난이도 있는 query를 생성할 수 있지만, 생성 과정에서 기존 문제에서 제공하는 information이 누락될 수 있다. 이러한 hallucination 현상 방지를 위하여 우리는 한정된 경우에서만 LLM prompting-based의 transformation 방법론을 사용하였다. Appendix에서 각 transformation의 detail 및 sample들을 기술한다.

% We employ two techniques to transform general QA reasoning tasks into binary decision tasks. 
We transform general QA reasoning tasks into binary decision tasks by modifying each sample into a \textit{positive} or \textit{negative} sample where the answer is ``Yes'' or ``No'', respectively.
In this section, we focus on the case of short-answer QA datasets.
We refer to Appendix \ref{sec:appendix:details_of_binary_decision_data_construction} for the multiple choice datasets.

For the positive sample, we perform a rule-based transformation on general queries by appending a confirmatory query. 
For example, the query "What is 1+1?" is transformed into "What is 1+1? Is the answer 2?".
This approach is straightforward for the cases where the correct answer is ``Yes''. 
Formally, given the question and its label, we use Prompt I to transform the prompt into the positive sample:

\begin{tcolorbox}[
    enhanced, 
    arc=0pt, 
    outer arc=0pt, 
    boxsep=0pt, 
    left=0pt, 
    right=0pt, 
    top=0pt, 
    bottom=0pt, 
    colback=darkgray, 
    colframe=darkgray
]
  % Upper sub-box with dark gray background and shorter height
  \begin{tcolorbox}[
      colback=darkgray, 
      colframe=darkgray, 
      coltext=white, 
      arc=0pt, 
      outer arc=0pt, 
      boxsep=5pt, 
      left=5pt, 
      right=5pt, 
      top=5pt, 
      bottom=-0.5pt, 
      height=0.35cm, % Set the desired height for the upper box
      valign=center, % Vertically center the content
      borderline south={0pt}{0pt}{darkgray}
  ]
    Prompt I (Positive Sample Transformation)
  \end{tcolorbox}
  % Lower sub-box with light gray background and vertically centered text
  \begin{tcolorbox}[
      colback=lightgray, 
      colframe=lightgray, 
      coltext=black, 
      arc=0pt, 
      outer arc=0pt, 
      boxsep=5pt, 
      left=5pt, 
      right=5pt, 
      top=1pt, 
      bottom=1pt, 
      height=1.5cm,
      valign=center, % Vertically center the content
      borderline north={0pt}{0pt}{lightgray}
  ]
  % \footnotesize
    % \texttt{{\char`\\}c} Q: {\char`\\}? Options: {\char`\\}o A:
    You are given a question and you MUST answer Yes or No.
    Question: $\{question\}$ Is the answer $\{label\}$? Answer:
  \end{tcolorbox}
\end{tcolorbox}

To transform into the negative sample, a wrong label is necessary based on the query and answer which will be placed in \textit{label} in Prompt I. 
For this, we obtain suitable incorrect answers by prompting GPT-4 (\citealp{openai-2023-gpt4}). 
% Subsequently, we use the aforementioned rule-based method to convert the original QA samples into binary decision samples where the answer is ``No''. 
We present GPT-4 prompts for the wrong label generation in Appendix \ref{sec:appendix:details_of_binary_decision_data_construction}.
% Secondly, we transform the original queries into natural binary decision queries through the prompting of the LLM. 
% This method can generate more realistic and challenging queries than rule-based transformation, but there is a risk of omitting information provided in the original problem, known as hallucination. 
% To prevent this issue, this LLM prompting-based transformation methodology is employed in limited cases. 
% The details and samples of each transformation method are described in the Appendix.
\input{figures/negative_head_}
\paragraph{Datasets and models}
We utilize StrategyQA, MuSiQue, subsets of MetaMATH (GSM8k-Rephrased and MATH-Rephrased), and AR-LSAT for evaluation, covering three domain tasks: multi-hop open-domain QA, mathematical reasoning, and logical reasoning. StrategyQA is a yes-no QA dataset, MuSiQue and the rephrased versions of GSM8k and MATH are short-answer QA datasets, and AR-LSAT is a multiple-choice dataset. 
Samples are in Table \ref{tab:appendix:example_of_datasets}.
To assess reasoning capabilities in binary decision tasks, we convert these datasets into binary decision sets using the techniques above, except for StrategyQA. 
AR-LSAT is converted to the yes-no QA formats, while the others are rephrased for answer verification formats. 
Details on the conversion can be found in Appendix \ref{sec:appendix:details_of_binary_decision_data_construction}.
We sample a portion of the entire dataset, and detailed statistics can be found in Appendix \ref{appen:data_statistics}. 
For the search space of LLMs, we explore LLaMA3-8B-Instruct (LLaMA), Mistral-7B-Instruct-v0.3 (Mistral), Gemma-1.1-7b-it (Gemma), Qwen2-7B-Instruct (Qwen), and GPT-4.

\subsubsection{Observation of Negative Bias}

% \paragraph{Precision and Recall}
% 우리는 또한 긍정과 부정의 response의 정확도를 비교하였다. 구체적으로 우리는 precision과 recall, ``Yes''에 대한 응답과 ``No''에 대한 응답 중 정답 비율, respectively을 측정하였다.
% 놀랍게도, Gemma in mathematical reasoning tasks를 제외한 모든 경우에서 precision이 recall에 비해 높았다.
% 이것은 모델의 negative bias으로 인해 답변 내용에 따른 정확도에 불균형이 있음을 의미한다.
% Fig \ref{fig:precision_recall}는 각 dataset들에 대한 model들의 precision과 recall 결과를 나타낸다. 대부분의 결과에서 precision에 비해 recall이 유의미하게 하락하는 것을 확인할 수 있는데, 이는 binary decision의 정답이 ``Yes''인 case에 대해 ``No''로 대답하는 경우가 증가함을 의미한다.  Fig \ref{fig:conf_hist}는 mathematical and logical reasoning task에서 LLaMA3 model 응답 confidence의 histogram 을 나타낸다. 전반적으로 negative response의 빈도가 positive에 비해 많으며, 더 강한 confidence를 가짐을 보여준다. 정리하자면, 우리는 statistical observation을 통해서 LLM들은 대체로 complex reasoning task에서 1) positive sample에 대한 정확도가 떨어지며 (low recall) 2) positive보다 더 많은 수의 negative response를 생성하고 3) 이러한 negative response는 더 높은 confidence를 보인다는 사실을 발견한다. 즉, large language model은 binary decision task 형태의 complex reasoning 과정에서 negative 위주로 응답하는 bias를 가지고 있다.

Fig \ref{fig:precision_recall} presents the precision and recall results on the transformed datasets. 
We use FP, FN, TP, and TN to denote false positive, false negative, true positive, and true negative, respectively.
In most cases, a significant drop in recall compared to precision is observed. 
% indicating an increase in instances where the answer to a binary decision with the correct response as ``Yes'' is incorrectly given as ``No''. 
According to the definition, a high precision implies that a significant proportion of the samples predicted as positive by the model (TP+FP) are true positives.
 % By definition, high precision indicates that a high ratio of samples where the model's responses are positive (TP+FP) are positive cases (TP). 
 Low recall means that among all the positive samples (TP+FN), the model rarely responds correctly with positive answers (TP). 
 The phenomenon where existing models exhibit high precision but low recall can be explained to that \textit{the model is overly cautious in outputting positive responses}. 
 Conversely, this implies that the model outputs negative responses indiscriminately, indicating that the trustworthiness of negative responses and positive responses is not balanced.
 
Fig \ref{fig:conf_hist} shows a histogram of the response confidence of the LLaMA in mathematical and logical reasoning tasks. 
% The X-axis denotes confidence for all lines. 
% Each histogram represents FP, FN, TP, and TN.
We observe that the overall frequency and confidence of negative responses (FN and TN) tend to be higher than those of positive responses (FP and TP). 
This indicates that \textit{the model tends to output negative responses more frequently and with greater confidence}.
% Generally, the frequency of negative responses is higher than that of positive responses, and these negative responses exhibit stronger confidence. 

In summary, we find that LLMs typically show 1) over-cautiousness to the positive response, and 2) more frequent and confident negative responses. 
It is evident that \textit{large language models possess a bias towards negative responses in the binary decision-making process of complex reasoning tasks}. 
A case study of the negative responses about the negative bias can be found in Appendix \ref{appendix:neg_res_analysis}. 

%% file: figures/negative_head_.tex
\begin{figure*}[t]
  \includegraphics[width=0.95\textwidth]{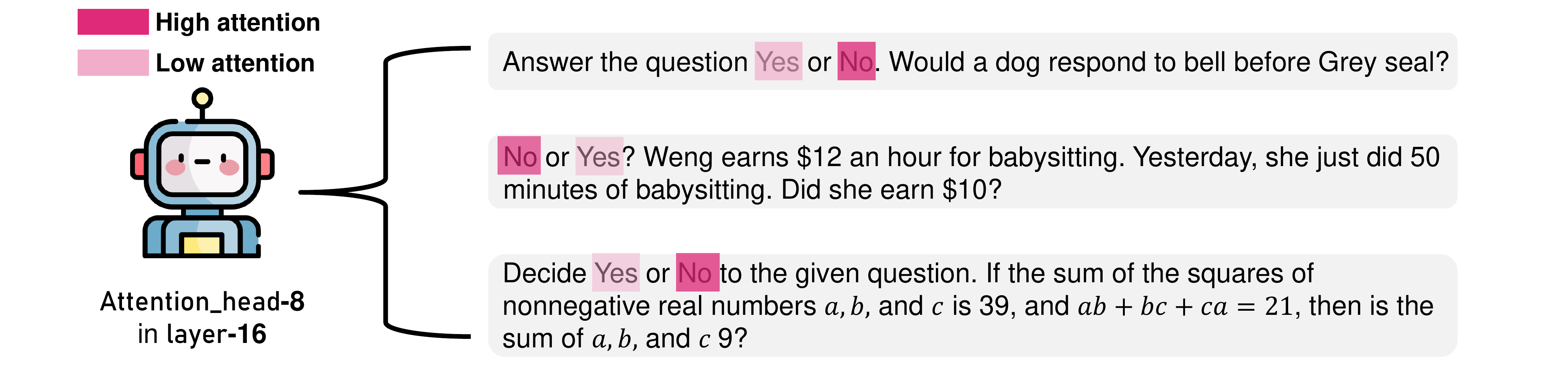}
  \caption{An example of the negative attention head. Three queries in the figure are sampled from StrategyQA (\citealp{geva2021strategyqa}), rephrased GSM8K and MATH datasets (\citealp{yu2023metamath}).}
  \label{fig:negative_head}
\end{figure*}

%% file: sections/3.2_Formulation_NAS.tex
\subsection{Formulation of Negative Bias} \label{3.2.formulation_nas}
\subsubsection{NAS: Negative Attention Score}

In this section, we formulate negative bias from the perspective of model intrinsic properties. 
Specifically, we focus on the model's internal attention patterns. 
To obtain an answer for a binary decision, a user generally provides answer candidates like ``Yes'' or ``No'' as instructions to the model before posing the query prompt (c.f., Prompt I). 
The model follows these instructions and responds to the given query in the manner provided in the instructions. 
During this response process, the model attends to the given instructions, and due to the operational characteristics of attention-based models, the candidate with the larger attention weight generally appears in the response. 
In this context, the negative bias of the LLM can be seen as originating from assigning greater attention weight to negative answer candidates during the reasoning process. 
Fig \ref{fig:negative_head} illustrates an example of negative attention head. 
To validate this rationale, we define Negative Attention Score (NAS) using the attention weights applied to negative answer candidate tokens like ``No'', and positive answer candidate tokens like ``Yes'', observed in a attention head.

% Formally, prompt의 길이를 $L_P$, instruction 내 ``Yes''와 ``No'' tokens의 index를 $t_{Yes}$와 $t_{No}$, instruction의 길이를 $L_I$라 하자.
% $l$번째 layer의 $h$번째 attention head에서 추론한 attention weight를 $A^{l,h}\in \mathbb{R}^{L_p \times L_P}$라고 했을 때, 의 NAS는 다음과 같다:
Let $L_x$ be the length of the sample $x$, $t_{Yes}$ and $t_{No}$ the positions of the ``Yes'' and ``No'' tokens within the instruction, and $L_I$ the length of the instruction in $x$. 
For the attention weight inferred by the $h$-th attention head in the $l$-th layer, denoted as $A^{l,h} \in \mathbb{R}^{L_x \times L_x}$, the NAS is defined as:
\begin{align}
    \text{NAS}^{l,h}_{x} &:= \notag\\
    & \hspace{-2.5em}\sum^{L_x}_{i=L_I}{\left(A^{l,h}_{i,t_{Yes}} + A^{l,h}_{i,t_{No}}\right) * \log\left(\frac{A^{l,h}_{i,t_{No}}}{A^{l,h}_{i,t_{Yes}}}\right)},\label{eq:m_cl}
\end{align}
where we omit the input $x$ in the attention function for brevity.
% $i$번째 row에서, NAS의 항은 두가지 factor로 consist되어있다. 
% 첫번째는 ``Yes''와 ``NO'' token에 가해지는 attention weight의 합으로, answer의 option에 해당하는 token에 높은 비중을 둔 attention head를 찾기 위함이다. 
% 두번째는 ``Yes''와 ``NO'' token에 가해지는 attention weight의 log 비율이다. 
% 이는 두 tokens 중 ``No'' token에 가해지는 attention weight가 ``Yes'' token의 그것 대비 높은 attention head를 찾기 위함이다.
The NAS term consists of two factors: the sum of the attention weights applied to the ``Yes'' and ``No'' tokens, which aims to find attention heads focusing significantly on the tokens representing answer options; and the logarithm of the ratio of attention weights applied to these tokens, identifying heads that preferentially attend to the ``No'' token over the ``Yes'' token.

NAS can have a value for each attention head on a single sample.
Based on this, we introduce two types of NAS variants: \textbf{single head NAS} and \textbf{model NAS}. 
Single head NAS represents the average NAS of a specific attention head across the sample(s). 
It is used to quantify the negative bias of a single attention head. 
On the other hand, model NAS represents the total NAS across all attention heads for the sample(s). 
This is used to approximate the model's negative bias.
Formally, given a sample set $X$, we define the single head NAS of the $h$-th attention head in the $l$-th layer and model NAS as follows:
% For example, we can formally express the single head NAS of $h$-th attention head in $l$-th layer on the dataset $X$ as $\frac{1}{|X|} \sum_{x_i \in X} \text{NAS}^{l,h}_{x_i}$.
\begin{align}
    \text{NAS}(X, l, h) := \frac{1}{|X|} \sum_{x_i \in X} \text{NAS}^{l,h}_{x_i}\textbf{ \footnotesize(single head)} \notag \\
    \text{NAS}(X, L, H) := \sum_{l \in L} \sum_{h \in H} \text{NAS}(X, l, h)\textbf{ \footnotesize (model)} \notag,
    \label{eq:single_and_model_nas}
\end{align}
where $L$ and $H$ are the sets of all layer and attention head indices, respectively.

\subsubsection{Empirical Studies}\label{subsubsec:empirical_studies}
% 앞서 정의된 NAS가 실제 negative bias에 대한 효과적인 지표임을 demonstrate하기 위해 우리는 model response에서 나타나는 negative confidence와 model NAS 사이의 correlation을 측정한다.
% 우리는 해당 coefficient 측정을 위하여 StrategyQA, rephrased GSM8K, AR-LSAT로부터 각각 500개의 sample을 추출하여 총 1500개로 구성된 test set을 구성한다. Table \ref{tab:nas_nconfs_corr}은 각 models에 대해 측정된 NAS와 negative response confidence 간의 Pearson correlation, Spearman's rank correlation coefficient을 보여준다. 모든 경우에서 NAS가 실제로 negative confidence와 positive한 상관관계가 있음을 확인할 수 있다. 이를 통해 NAS가 model의 negative bias를 나타내는 효과적인 지표라고 볼 수 있다. 

\input{tables/nas_nconfs_correlation}

\textbf{NAS and Negative Confidence} To demonstrate that the previously defined NAS is an effective indicator of negative bias, we measure the correlation between model NAS and the negative confidence observed in model responses. 
For this measurement, we construct a test set consisting of a total of 1,500 samples, drawing 500 samples each from StrategyQA, rephrased GSM8K, and rephrased AR-LSAT.
Table \ref{tab:nas_nconfs_corr} shows the Pearson correlation and Spearman's rank correlation coefficient measured between model NAS and negative response confidence for each model. 
In all cases, we confirm that model NAS has a positive correlation with negative confidence. 
This validates that NAS is an effective indicator of a model's negative bias.

% 추가적으로, 우리는 NAS를 통해 정의되는 negative heads들에 대한 study를 진행한다. 앞서 생성한 3가지 domain으로 이루어진 QA dataset에 대해서, 우리는 각각의 domain subset을 기준으로 3가지 경우에서 negative attention heads를 탐색한다. Subset 내부의 각 sample 별로 측정한 single head NAS가 가장 큰 top-200 heads들 중 subset 전체 subset sample의 90\% 이상에서 공통적으로 나타나는 heads들을 해당 domain subset에 대한 negative attention head로 지칭한다. 우리는 model의 종류를 고정한 상태로 각 subset에서 이러한 negative heads들을 추출한 뒤에 이들 세 그룹이 overlap되는 정도를 측정하였다. Table \ref{tab:neg_heads_overlap}에서 overlapping 비율을 나타내고 있는데, 놀랍게도 서로 다른 domain subset을 통해 추출한 negative heads가 높은 비율로 overlap되는 것을 확인할 수 있다. 이를 통해서 우리는 query-agnostic하게 높은 NAS를 나타내는, 즉 model의 negative bias의 공통적 원인이 되는 negative attention head가 존재함을 알 수 있다.
\input{figures/nasa_}

\input{tables/negative_head_overlaps}
\textbf{Overlapping Negative Attention Heads} We conduct a study on negative attention heads defined through NAS. 
Using our previously constructed test set, which spans three different domains, we investigate whether the negative attention heads extracted from each domain subset overlap with one another.
For each sample $x_i$ in a domain subset $X$, we first define the tuple list $P^i$ consisting of the layer and head indices of the top-$k$ attention heads with the highest single head NAS:
\begin{align}
P^i_X &= \text{Top-}k_{(l, h)} \, \text{NAS}(\{x_i\}, l, h)\text{.}
\end{align}
Among these extracted heads, we select those that are consistently included in over 90\% of the tuple lists, which we denote as $C_X$:
% \begin{align}
% C_X &= \{ (l, h)\mid \notag \\
% &(l, h) \text{ appears in at least } 0.9 \times |X| \text{ of } P^i_X \}\text{.}
% \end{align}
\begin{align}
&C_X = \left\{ (l,h) \ \middle| \right. \notag \\
&\left. (l, h) \text{ appears in at least } 0.9 \times |X| \text{ of } P_X^i \right\}.
\end{align}
Finally, we extract the list of top $N$ attention heads $P_X$, sorted by their single head NAS values on $X$. 
\begin{align}
P_X &= \text{Top-}N_{(l, h) \in C} \, \text{NAS}(X, l, h)\text{.}
\end{align}
In this process, we set $k$ to 200 and $N$ to 100.

% Specifically, we identify negative attention heads for a domain subset as those among the top-200 heads with the highest single head NAS per sample that appear in over 90\% of the subset's samples.
Keeping the model type constant, we extract such negative attention heads from each domain subset and then measure the extent of their overlap.
The overlapping rates are displayed in Table \ref{tab:neg_heads_overlap}, and surprisingly, we find that negative heads extracted through different domain subsets significantly overlap. 
This indicates the existence of \textit{query-agnostic negative attention heads} that represent a common underlying cause of the model's negative bias.

%% file: tables/nas_nconfs_correlation.tex
\begin{table}[t]
    \centering
    \caption{Correlations of NAS and the negative confidence ($-1$ to $1$).}
    {\resizebox{0.47\textwidth}{!}{
    \renewcommand{\arraystretch}{1.1}    
    \begin{tabular}{l|cccc}
    \toprule
      \textbf{Model} &\textbf{LLaMA} & \textbf{Mistral}  & \textbf{Gemma} & \textbf{Qwen} \\
     \hline
    \textbf{Pearson} & 0.5274& 0.3625&0.6322& 0.4806
\\
    \textbf{Spearman} & 0.5201& 0.3563&0.6725& 0.5166\\
     \bottomrule
    \end{tabular}
    }}
    \label{tab:nas_nconfs_corr}
\end{table}

% \begin{wraptable}{r}{0.5\textwidth}
%     \centering
%     \caption{Pearson correlations of NAS and the negative confidence ($-1$ to $1$).}
%     {\resizebox{0.5\textwidth}{!}{
%     \renewcommand{\arraystretch}{1.1}    
%     \begin{tabular}{cccc}
%     \toprule
%       \textbf{LLaMA3-8B} & \textbf{Mistral-7B}  & \textbf{Gemma-7B} & \textbf{Qwen2-7B} \\
%      \hline
%     0.53& 0.36&0.63& 0.48
% \\
%      \bottomrule
%     \end{tabular}
%     }}
%     \label{tab:nas_nconfs_corr}
% \end{wraptable}

%% file: figures/nasa_.tex
\begin{figure*}[!t]
  \includegraphics[width=\textwidth]{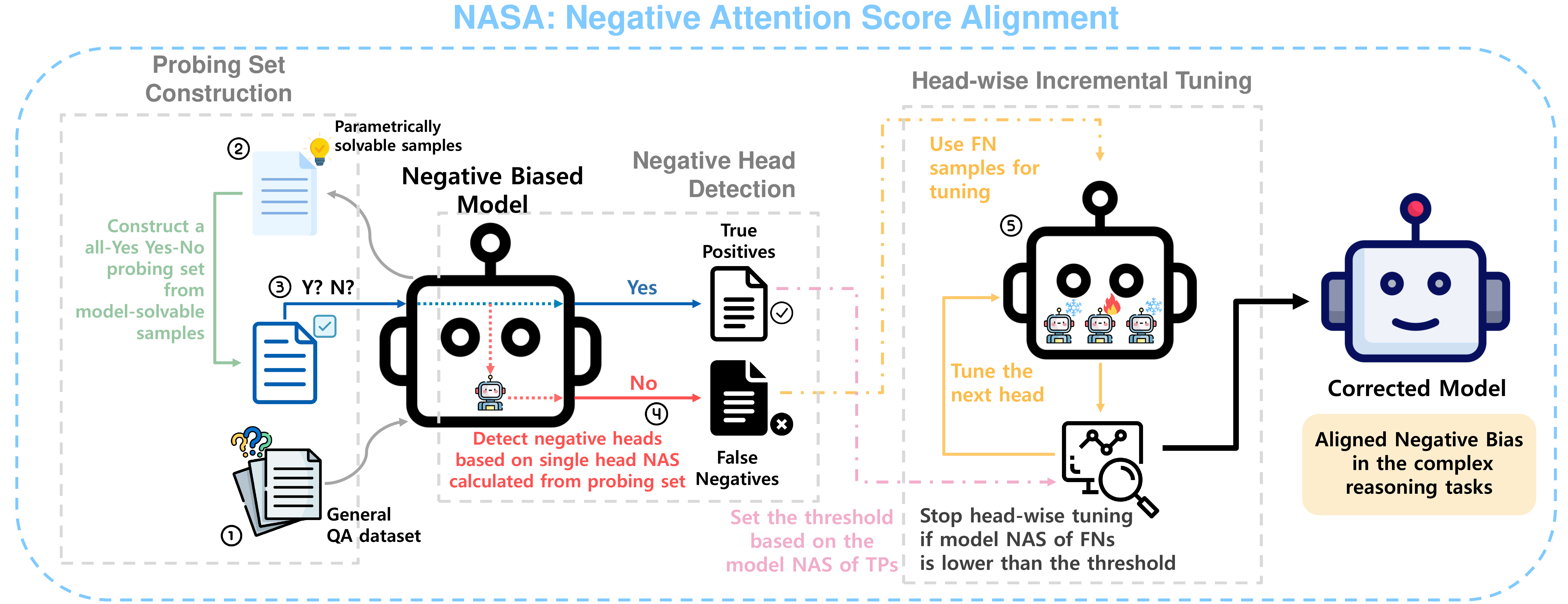}
  \caption{Overall framework of negative attention score alignment method.}
  \label{fig:nasa}
\end{figure*}

%% file: tables/negative_head_overlaps.tex
\begin{table}[t]
    \centering
    \caption{Overlapping proportion of top-100 negative heads for various datasets ($0$ to $1$).}
     {\resizebox{0.35\textwidth}{!}{
    \begin{tabular}{cccc}
    \toprule
      \textbf{LLaMA} & \textbf{Mistral}  & \textbf{Gemma} & \textbf{Qwen} \\
     \midrule
       0.74  & 0.76 & 0.80 & 0.74\\
     \bottomrule
    \end{tabular}
    }}
    \label{tab:neg_heads_overlap}
\end{table}

% \begin{wraptable}{r}{0.5\textwidth}
%     \centering
%     \caption{Overlapping proportion of top-200 negative heads for various datasets ($0$ to $1$).}
%      {\resizebox{0.5\textwidth}{!}{
%     \begin{tabular}{cccc}
%     \toprule
%       \textbf{LLaMA3-8B} & \textbf{Mistral-7B}  & \textbf{Gemma-7B} & \textbf{Qwen2-7B} \\
%      \midrule
%        0.74  & 0.76 & 0.80 & 0.74\\
%      \bottomrule
%     \end{tabular}
%     }}
%     \label{tab:neg_heads_overlap}
% \end{wraptable}

%% file: sections/4_Method.tex
\section{NASA: NAS Alignment Framework} \label{4.method}

Motivated by previous observations, we propose \textbf{NAS} \textbf{A}lignment (\textbf{NASA}), a framework designed to address attention heads that induce negative bias in a parameter-efficient manner. 
In NASA, we start by constructing a probing set to identify negative attention heads, followed by a selection process to determine which of these heads will be fine-tuned. 
Our framework fine-tunes the target attention heads in the order based on single head NAS.

\subsection{Probing Set Construction}\label{subsec:probing_set_construction}

% To observe the negative bias in attention heads, we start by constructing a negative attention head probing set. 
To construct the probing set, we use a short-answer QA dataset and select 
the parametric samples where the model can correctly answer the question.
A detailed explanation of the parametric sample selection can be found in Appendix  \ref{sec:appendix:details_of_parametric_sample_selection}.
Next, we convert the selected parametric samples into binary decision making format using Prompt I in Section \ref{subsubsec:binary_decision}.
Note that in the probing set, the label is uniformly set to ``Yes''.
% to effectively reveal the model's negative bias.
The probing set is designed to have two properties: i) the model possesses knowledge about the question, and ii) the model should respond ``Yes'' to the converted question.
We assume that attention heads with a high single head NAS for the converted samples that meet the two conditions are strongly associated with negative bias and should be addressed.

In Section \ref{subsubsec:empirical_studies}, we observe that the negative bias attention heads exhibit query-agnostic properties. 
Based on this observation, we employ another multi-step reasoning dataset, HotpotQA (\citealp{yang-etal-2018-hotpotqa}), to construct the probing set. 
We presume that HotpotQA, being less challenging compared to recent datasets, has a higher proportion of samples that can be answered using the parametric knowledge of LLMs.
% For the samples selected for probing, we construct the probing set by converting them into the form of binary decision prompt (See Section \ref{subsubsec:binary_decision}).

\subsection{Negative Attention Head Probing}\label{subsec:attention_head_probing}
% We probe negatively biased attention heads using the constructed probing set.
Using the negative attention head detection process described in Section \ref{subsubsec:empirical_studies}, we extract the attention heads to be fine-tuned by utilizing the probing set $X$.
In this process, we set $k$ to 100 and $N$ to 30.
% Old version, see 3.2.2
% First, for each sample $x_i$ in the probing set $X$, we measure the single head NAS of every attention head. 
% For each sample $x_i$, we define a tuple list $P^i$ consisting of the layer indices and head indices of the top-$k$ attention heads with the highest NAS scores, as follows:
% \begin{align}
% P^i &= \text{Top-}k_{(l, h)} \, \text{NAS}(\{x_i\}, l, h)\text{.}
% \end{align}

% Among these extracted heads, we select those that are consistently included in over 90\% of the tuple lists, which we denote as $C$:
% \begin{align}
% C &= \{ (l, h)\mid \notag \\
% &(l, h) \text{ appears in at least } 0.9 \times |X| \text{ of } P^i \}\text{.}
% \end{align}
% Finally, the list of top $N$ attention heads $P$, sorted by their single head NAS values on $X$, are targets for fine-tuning. 

% \begin{align}
% P &= \text{Top-}N_{(l, h) \in C} \, \text{NAS}(X, l, h)\text{.}
% \end{align}
% In this paper, we set $k$ to 100 and $N$ to 30.

At this point, we further categorize the probing set $X$ into cases where the model correctly answers $TP(X)$ (i.e., the true positive set) and cases where it does not $FN(X)$ (i.e., the false negative set). 
During the fine-tuning phase of the model (See Section \ref{subsec:finetuning_for_debiasing}), we use the false negative set as the fine-tuning set $D$. 
This approach is aimed at inducing the model to generate positive answers for samples that it fails to answer due to a negative bias. 
Meanwhile, the true positive set is used to determine the threshold $\tau$ for early halting of the fine-tuning process. 
\begin{align}
    \tau := \min_{x \in TP(X)} \text{NAS}(\{x\}, L, H)
    \label{eq:early_halting_threshold}
\end{align}
In other words, during the training process aimed at reducing the model NAS, if the model NAS on the validation set (i.e., subset of $FN(X)$) falls below the minimum model NAS of true positive set, training is halted to minimize the bias towards positives.

\subsection{Head-wise Incremental Tuning}\label{subsec:finetuning_for_debiasing}
% 본 섹션에서, 우리는 모델 내 the bias를 발현시키는 attention head를 detection하고 (Section 4.2) 모델의 spefine-tuning을 통해 
% Section 4.3에서 specify한 attention heads를 addressing하기 위해, 우리는 additional supervised fine-tuning을 진행한다.
% 우리의 fine-tuning method는 negative bias attention heads의 query and key projection weight만 fine-tuning 하기 때문에 parameter efficient하다.
% To address the identified attention heads as specified in Section \ref{subsec:attention_head_probing},  we propose an additional supervised fine-tuning method. 
% Our fine-tuning method is parameter efficient as it focuses only on fine-tuning the query and key projection weights of the selected negative bias attention heads.

\paragraph{Training setup} 
% 우리는 attention head probing에 사용된 data를 fine-tuning에 활용한다.
% 모델은 question을 맞추기 위한 parametric knowledge를 보유하고 있기 때문에 우리는 prompt II에 대해서 positive answer를 return하도록 fine-tuning data를 구성한다.
% Probing set은 Condition I을 충족하기 때문에 우리는 related facts가 포함된 context를 모델에게 제공하지 않는다.
As mentioned in Section \ref{subsec:attention_head_probing}, we utilize the $FN$s in the probing set for fine-tuning. 
Because the model fails to generate positive answers for the samples although the model possesses parametric knowledge, it is expected that fine-tuning with these samples can effectively address the negative bias.
We do not provide contexts containing related facts to the model because the model already has the corresponding knowledge.

\input{tables/main_result_multihop}
\input{tables/main_result_math}
Similar to a standard supervised fine-tuning objective, we target only the answer token (e.g., ``Yes'') for training. 
Since the fine-tuning set contains only short answers, training the model to generate the end of a sequence token could introduce a bias towards short responses. 
To prevent this side effect, we exclude the end of a sequence token in a loss calculation.

\paragraph{Training strategy}
% 우리는 attention head probing 단계에서 평균 NAS가 높았던 attention head 순서로 하나씩 fine-tuning한다.
% 우리는 target answer가 positive인 data로 학습하기 때문에 positive bias를 유발할 위험성이 있다.
% 이를 prevent하기 위해 우리는 dataset의 일부를 validation set으로 두어, early stopping scheme을 적용한다.
% Early stopping criteria는 validation set에 대한 모델 전체 attention head의 NAS 합과 fine-tuning 대상 attention head의 NAS 값이다.
% 전자의 경우 매 epoch 마다 값이 감소하는지 확인하고, 증가하는 경우 학습을 중단한다.
% 또한 후자의 경우 매 epoch 마다 값이 감소하는지와 특정 threshold 이하의 값 이상을 유지하는지를 확인한다.
% 두 조건 중 하나라도 충족하지 못하는 경우 역시 학습을 중단한다.
We fine-tune the query and key projection modules of each attention head sequentially in order of decreasing single head NAS which is measured during the probing stage. 
% This procedure is parameter-efficient as it focuses only on partial weights of the selected negative bias attention heads.

Since we fine-tune with data where the target answer is positive, there is a risk of inducing a positive bias.
Furthermore, updating a single attention head might change the NAS of another attention head.
To prevent these, we set aside a portion of the dataset as a validation set and apply \textit{early stopping} and \textit{update cancellation} schemes for each attention head tuning and \textit{early halting} scheme for the whole fine-tuning process.
% Algorithm for the entire process can be found in Appendix \ref{appendix:nasa_algorithm}.

% \paragraph{Strategy of attention head tuning subprocess}
For early stopping, we set criteria based on the model NAS and single head NAS.
For the former criteria, we check for a decrease in the value after each epoch.
For the latter one, we verify that i) the value decreases and ii) remains above a specific threshold $\rho$ after each epoch.
Fine-tuning of the current attention head is stopped if either of the three conditions is not met.

For the update cancellation, after fine-tuning each attention head, we calculate the difference between the single head NAS and the model NAS before and after fine-tuning. 
If any of these values have increased compared to before training, we revert the parameter update of that attention head to exclude attention heads that reinforce the negative bias from the training target.

% \paragraph{Strategy of the whole tuning process}
% 우리는 Section \ref{subsec:attention_head_probing}에서 언급한 바와 같이, true positive probing set 내의 샘플들의 NAS를 측정하여 최소값을 early halting threshold $\tau$로 설정한다. 
% 우리는 hypothesize한다 false negative sample에서의 NAS 값이 true positive에서의 그것보다 낮아지는 경우 positive bias를 초래할 수 있다고. 
% 각 negative bias attention head의 학습 완료 후, 우리는 validation set에서 전체 attention head에서의 NAS를 측정한다.
% 이것이 $\tau$보다 작아지는 경우 학습은 중단되어 local optima에 빠지는 것을 방지한다.
For the early halting, we utilize the model NAS of each sample within the $TP(X)$, $\tau$, as mentioned in Section \ref{subsec:attention_head_probing}. 
This threshold setting is based on the rationale that if the model NAS in a $FN(X)$ is lower than in a $TP(X)$, it may induce a positive bias. 
After training each negative attention head, we measure the model NAS on the validation set. 
If this value falls below $\tau$, training is halted to prevent falling into local optima.
Algorithm for the entire process can be found in Appendix \ref{appendix:nasa_algorithm}.

%% file: tables/main_result_multihop.tex
\begin{table*}[!t]
\centering
\caption{Model evaluation results on multi-step reasoning datasets.}
{\resizebox{0.95\textwidth}{!}{
\renewcommand{\arraystretch}{1.1}
\footnotesize
\begin{tabular}{@{}l|c||ccccc@{}}
\toprule
\textbf{Dataset} & \textbf{Model}                        & \textbf{Accuracy}   & \textbf{Precision} & \textbf{Recall}    & \textbf{F1}     & \textbf{NAS} \\ 
\hline
\multirow{3}{*}{StrategyQA} & LLaMA / + NASA           & 0.871 / \textbf{0.877}       & \textbf{0.919} / 0.864     & 0.795 / \textbf{0.875}     & 0.852 / \textbf{0.869}  & 160.6 / \textbf{55.0}  \\
&Mistral / + NASA       & \textbf{0.842} / 0.830      & \textbf{0.882} / 0.802    & 0.772 / \textbf{0.861}     & 0.823 / \textbf{0.830}  & 114.1 / \textbf{34.3}  \\
&Gemma / + NASA                & \textbf{0.778} / 0.771     & \textbf{0.923} / 0.907    & 0.676 / \textbf{0.690}    & 0.777 / \textbf{0.784} & 107.8 / \textbf{71.8}  \\
&Qwen / + NASA                & 0.829 / \textbf{0.844}     & \textbf{0.921} / 0.907    & 0.694 / \textbf{0.742}   & 0.792 / \textbf{0.816} & 209.6 / \textbf{107.2}  \\
% \cline{2-7}
% &Ours (LLaMA 3)                  & \textbf{0.766}      & 0.809     & \textbf{0.653}     & \textbf{0.723}  & -      \\
% &Ours (Mistral)                & 0.636      & 0.609     & \textbf{0.843}     & \textbf{0.707}  & -      \\
% &Ours (Gemma)                  & \textbf{0.625}      & 0.847     & 0.243     & 0.377  & -      \\
\hline
\multirow{3}{*}{\parbox{1.7cm}{MuSiQue\\(Rephrased)}}   &LLaMA / + NASA       & 0.720 / \textbf{0.757}     & \textbf{0.859} / 0.815     & 0.555 / \textbf{0.683}    & 0.675 / \textbf{0.743} & 199.8 / \textbf{64.0} \\
& Mistral / + NASA       & 0.649 / \textbf{0.707}     & \textbf{0.848} / 0.802    & 0.416 / \textbf{0.594}    & 0.558 / \textbf{0.682} & 131.2 / \textbf{110.7}  \\
& Gemma / + NASA                & \textbf{0.518} / 0.503     & \textbf{0.837} / 0.834     & 0.529 / \textbf{0.555}    & 0.648 / \textbf{0.667} & 259.8 / \textbf{96.2}  \\
&Qwen / + NASA                & 0.621 / \textbf{0.639}     & \textbf{0.883} / 0.858    & 0.482 / \textbf{0.552}   & 0.624 / \textbf{0.672} & 418.8 / \textbf{282.9}  \\
% \cline{2-7}
% &Ours (LLaMA 3)                  & \textbf{0.745}     &  0.848     & \textbf{0.597}     & \textbf{0.701}  & -      \\
% &Ours (Mistral)                & \textbf{0.693}      & 0.694     & \textbf{0.706}     & \textbf{0.700}  & -      \\
% &Ours (Gemma)                  & \textbf{0.625}      & \textbf{0.821}     & 0.517     & 0.634  & -      \\
% \hline
% \multirow{3}{*}{2WikiMultiHopQA} & LLaMA3-8B / + NASA            & 0.745 / 0.727     & 0.746 / 0.685    & 0.744 / \textbf{0.840}    & 0.745 / \textbf{0.755}  & 201.8 / - \\
% &Mistral-7B / + NASA       & 0.663 / 0.644     & 0.667 / 0.603     & 0.661 / \textbf{0.867}    & 0.664 / \textbf{0.711} & 310.3 / -  \\
% &Gemma-7B / + NASA                & 0.584 / \textbf{0.681}     & 0.719 / 0.678     & 0.639 / \textbf{0.690}    & 0.676 / \textbf{0.683}  & 129.3 / -  \\
% &Qwen2-7B / + NASA                & - / -     & - / -    & - / -   & - / - & - / -  \\
% \cline{2-7}
% &Ours (LLaMA 3)                  & 0.727      & 0.685     & \textbf{0.840}     & \textbf{0.755}  & -      \\
% &Ours (Mistral)                & 0.644      & 0.603     & \textbf{0.867}     & \textbf{0.711}  & -      \\
% &Ours (Gemma)                  & \textbf{0.681}      & 0.678     & \textbf{0.690}     & \textbf{0.683}  & -      \\
\bottomrule
\end{tabular}
}}
\label{tab:main:multihop}
\end{table*}

%% file: tables/main_result_math.tex
\begin{table*}[!t]
\centering
\caption{Model evaluation results on mathematical and logical reasoning datasets.}
{\resizebox{0.95\textwidth}{!}{
\renewcommand{\arraystretch}{1.1}
\footnotesize
\begin{tabular}{@{}l|c||ccccc@{}}
\toprule
\textbf{Dataset} & \textbf{Model }                        & \textbf{Accuracy}   & \textbf{Precision} & \textbf{Recall}    & \textbf{F1}     & \textbf{NAS} \\ 
\hline
\multirow{3}{*}{\parbox{1.7cm}{GSM8K\\(Rephrased)}}& LLaMA / + NASA           & 0.537 / \textbf{0.545}     & \textbf{0.587} / 0.531    & 0.336 / \textbf{0.838}    & 0.428 / \textbf{0.650}  & 218.2 / \textbf{74.7} \\
&Mistral / + NASA       & 0.501 / \textbf{0.512}     & \textbf{0.518} / 0.509    & 0.163 / \textbf{0.782}    & 0.247 / \textbf{0.617} & 115.5 / \textbf{77.0} \\
&Gemma / + NASA                & \textbf{0.533} / 0.524     & \textbf{0.546} / 0.541    & 0.837 / \textbf{0.887}    & 0.661 / \textbf{0.672} & 159.6 / \textbf{66.2} \\
&Qwen / + NASA                & 0.538 / \textbf{0.576}     & \textbf{0.640} / 0.614    & 0.534 / \textbf{0.738}   & 0.582 / \textbf{0.670} & 210.9 / \textbf{82.9}  \\
% \cline{2-7}
% &Ours (LLaMA 3)                  & 0.544& 0.542& 0.812& 0.650& -      \\
% &Ours (Mistral)                & -      & -     & -     & -  & -      \\
% &Ours (Gemma)                  & 0.523& 0.536& 0.885& 0.668& -      \\
\hline
\multirow{3}{*}{\parbox{1.7cm}{MATH\\(Rephrased)}}   &LLaMA / + NASA            & \textbf{0.563} / 0.549     & \textbf{0.587} / 0.534    & 0.530 / \textbf{0.874}     & 0.557 / \textbf{0.663}  & 183.3 / \textbf{60.8}  \\
& Mistral / + NASA       & 0.504 / \textbf{0.526}     & \textbf{0.527} / 0.522     & 0.136 / \textbf{0.685}    & 0.216 / \textbf{0.592}  & 117.9 / \textbf{109.6} \\
&Gemma / + NASA                & \textbf{0.546} / 0.543     & \textbf{0.556} / 0.553    & 0.810 / \textbf{0.871}    & 0.659 / \textbf{0.677} & 185.0 / \textbf{64.0} \\
&Qwen / + NASA                & 0.564 / \textbf{0.581}     & \textbf{0.661} / 0.617    & 0.537 / \textbf{0.715}   & 0.593 / \textbf{0.662} & 224.0 / \textbf{103.1}  \\
% \cline{2-7}
% &Ours (LLaMA 3)                  & 0.552&  0.534& 0.766& 0.629& -      \\
% &Ours (Mistral)                & -     &  -     & -     & -  & -      \\
% &Ours (Gemma)                  & -     &  -     & -     & -  & -      \\
\hline
\multirow{3}{*}{\parbox{1.7cm}{AR-LSAT\\(Rephrased)}}   & LLaMA / + NASA            & 0.513 / \textbf{0.518}     & \textbf{0.547} / 0.537    & 0.297 / \textbf{0.444}   & 0.385 / \textbf{0.486} & 178.1 / \textbf{63.6}  \\
& Mistral / + NASA       & 0.511 / \textbf{0.530}     & 0.547 / \textbf{0.553}     & 0.280 / \textbf{0.447}    & 0.370 / \textbf{0.494}  & 108.7 / \textbf{80.02} \\
& Gemma / + NASA                & 0.509 / \textbf{0.510}     & \textbf{0.525} / 0.523    & 0.468 / \textbf{0.534}    & 0.495 / \textbf{0.528} & 151.7 / \textbf{68.0} \\
&Qwen / + NASA                & 0.492 / \textbf{0.496}     & \textbf{0.529} / 0.528    & 0.110 / \textbf{0.182}   & 0.182 / \textbf{0.270} & 317.6 / \textbf{207.3}  \\
\bottomrule
\end{tabular}
}}
\label{tab:main_math}
\end{table*}

%% file: sections/5_Experiment.tex
\section{Experiments} \label{5.experiments}

\subsection{Experimental Setup}
% 우리는 LLaMA 3, Mistral, 그리고 Gemma를 실험에 사용했다.
% 우리는 Section 3에서 사용한 yes-no question answering dataset에 대해 evaluation하였으며, accuracy, precision, recall, F1, 그리고 전체 attention head의 probing set에 대한 NAS 값의 합을 측정하였다.

% For early stopping, 우리는 두개의 NAS 값 (See the last paragraph in Section 4.3)이 감소하지 않거나 target attention head의 validation set에 대한 NAS가 1.0 이하로 내려가면 학습을 중단하였다.
% Initial learning rate는 $1e-6$, batch size는 32, maximum number of epochs는 30으로 설정했다.
% 우리는 Alpaca repository\footnote[1]{\url{https://github.com/tatsu-lab/stanford_alpaca}} 를 참조하여, weight decay는 0, warmup ratio는 0.03으로 설정했다.
We use LLaMA, Mistral, Gemma, and Qwen for our experiments.
We use those supervised fine-tuned models before NASA-tuning as baselines.
We use LLaMA-Factory (\citealp{zheng2024llamafactory})\footnote[1]{\url{https://github.com/hiyouga/LLaMA-Factory}} and HuggingFace Transformers (\citealp{wolf-etal-2020-transformers})\footnote[2]{\url{https://github.com/huggingface/transformers}} for all the experiments.
The models are evaluated using binary decision datasets, which are described in Section 3, measuring accuracy, precision, recall, F1, and the model NAS. 
We set $\rho$ to 0.5, the initial learning rate to 1e$-$6, the batch size to 32, and the maximum number of epochs to 30. 
We reference the Alpaca repository\footnote[3]{\url{https://github.com/tatsu-lab/stanford_alpaca}}, setting the weight decay to 0 and the warmup ratio to 0.03.

\subsection{Main Results}\label{subsec:main_results}
% 우리는 Gemma를 제외한 대부분의 case에서 F1 score가 크게 증가하는 것을 관측한다. 
% 특히 우리의 방법론은 negative bias를 완화함에 따라 recall를 효과적으로 향상시킨다. 

% Note that Gemma의 the probing set에 대한 NAS 값은 다른 두 LLM baseline에 비해 낮다.
% 이는 즉 attention score에 기반했을 때 
% Gemma가 일부 dataset에서의 F1 score가 오히려 감소하는 것을 고려했을 때, NAS는 our method의 효과를 가늠할 수 있는 척도가 될 수 있다.

% 다른 데이터셋과 비교했을 때, 2WikiMultiHopQA의 NAS 값은 상대적으로 높으며, our method는 일관적으로 높은 F1 score 향상을 가져다 주었다.
% We observe a significant increase in the F1 score in all the cases.
% In particular, our method effectively enhances recall which implies that our approach mitigates the negative bias.
% Accuracy 또한 baseline을 surpass하는 경우가 있으며, 특히 Mistral-7B의 accuracy가 5\% 이상 상승했다 MuSiQue dataset에서.
% 우리는 확인할 수 있다 ours 중 NAS가 낮은 모델은 baseline에 비해 accuracy가 slightly 낮은 것을.
% 이는 의미한다, NAS는 prediction에 대한 bias와 높은 correlation이 있으며, 최소화하기 보다는 적정 수준을 유지하는 것이 필요함을. 
% In this section, 우리는 section \ref{3.problem}에서 관측하고 분석한 요소들을 기준으로 negative bias가 NASA에 의해 실제로 mitigate되었는지 확인한다. 
% Tables \ref{tab:main:multihop} and \ref{tab:main_math} show the performances of our method in various binary decision tasks. 일반적으로 accuracy가 향상되었으며, precision과 recall 간의 gap이 유의미하게 감소하였음을 관측할 수 있다. Section \ref{3.problem}에서 언급한 바와 같이 precision이 high하고 recall이 low하다는 것은 model이 positive response를 provide하는 것에 지나치게 신중함을 의미하며, 이는 바꿔말해서 negative response에 대한 trustworthiness의 degradation으로 이어진다.
% We examine the effectiveness of NASA in mitigating negative bias.
Tables \ref{tab:main:multihop} and \ref{tab:main_math} show the performance of our method across various binary decision tasks. 
Generally, we observe an improvement in accuracy and a significant reduction in the precision-recall gap.
As noted in Section \ref{3.problem}, high precision coupled with low recall indicates that the model is excessively cautious in providing positive responses, resulting in a degradation in the trustworthiness of negative responses. From this perspective, the negative bias can be mitigated by reducing the gap between precision and recall while maintaining accuracy, which balances the trustworthiness of the model's responses and ultimately improves the model.

Additionally, based on the improvement in the F1 score, which is the harmonic mean of precision and recall, we demonstrate that the balance between precision and recall has been achieved in the direction of improving the model's capability for reasoning. 
We also observe that our method effectively reduces the model NAS. 
In Appendix \ref{appen:ablation_study}, we demonstrate that the components of our method contribute effectively to tackling negative bias.

Meanwhile, the slight decrease in precision might superficially appear as a performance decline. 
However, it can be interpreted as enhancing the model's trustworthiness by adjusting the alignment between the model's actual knowledge and its responses. 
This means that 
precision might slightly decrease because some samples incorrectly categorized as true negatives due to negative bias might shift to false positives. Specifically, 
even if the model makes incorrect inferences or remains uncertain about a question, in a binary decision scenario, negative bias could lead to it being classified as a true negative. 
An analysis of the shifts in negative responses due to NASA can be found in Appendix \ref{appendix:res_shift_ratio_analysis}.

% We can observe that models with low NAS in our study show slightly lower accuracy compared to the baseline.
% This suggests that NAS is highly correlated with bias in predictions and that maintaining an appropriate level rather than minimizing it is necessary.

%% file: sections/6_Analysis.tex
\section{Analysis}\label{6.analysis}
% In this section, we analyze the impact of NASA on the generalization ability of LLMs beyond the binary decision task in which this work focus. After that, 우리는 NASA가 negative bias를 완화시키면서 model calibration 측면에 끼치는 영향을 분석한다. 또한, few-shot prompting 상황에서 few-shot example과 NASA의 성능 향상 측면에서의 관련성을 분석한다.
We analyze the impact of NASA on the general reasoning capability of LLMs beyond the binary decision task which is the primary focus of this work. After that, we examine the effect of NASA on model calibration. Additionally, regarding performance improvement and negative bias, we investigate the relationship between few-shot examples and NASA in few-shot prompting scenarios.
We present an instruction following analysis and an ablation study in Appendices \ref{appendix:generalization_to_universal_binary_decision}, \ref{appendix:transferrability_across_various_instructions} and \ref{appen:ablation_study}.

\input{sections/6.1_Preservation_of_General_Reasoning_Abilities}

% \input{sections/6.2_Generalization_to_Universal_Binary_Decision}

% \input{sections/6.3_Transferability_across_Various_Instructions}

\input{sections/6.4_Enhanced_Model_Calibration}
\input{tables/general_reasoning}
\input{tables/calibration_test}
% % \input{sections/6.4_Impact_of_CoT_to_Negative_Bias}

% % \input{sections/6.5_Comprehensive_Eval_with_GPT4}

% % \input{sections/6.6_Negative_Bias_in_Foundation_Models}

% % \input{sections/6.7_Ablation_Study}

\input{sections/6.3_Analysis_on_few_shot_scenario}

%% file: sections/6.1_Preservation_of_General_Reasoning_Abilities.tex
\subsection{General Reasoning Abilities}
% \input{tables/general_reasoning}
% In this section, 우리는 NASA가 적용된 large language model의 general QA reasoning에 대한 성능을 평가한다. 우리는 LLaMA3-8B와 해당 model에 NASA가 적용되었을 때의 MuSiQue, GSM8K, AR-LSAT에 대한 general reasoning 성능을 측정한다. 우리는 각 benchmark에서 model이 생성한 prediction에 answer가 포함되어 있는지 확인한 뒤에, GPT-4를 통해 answer가 포함되어 있는 prediction들이 실제로 answer와 같은 의미인지 재차 확인한다. 즉, GPT-4를 utilize하는 filtering 과정을 통해 model이 실제로 정답을 올바르게 예측하는 비율을 평가 metric으로서 측정한다. Table \ref{tab:general_reasoning}에서 보여지는 바와 같이, NASA가 적용된 model은 negative bias가 제거됨과 동시에 기존의 general reasoning ability가 유지되거나 더 향상되는 결과를 보여준다.

We evaluate the performance of LLMs fine-tuned with NASA on general QA reasoning tasks. 
We utilize general reasoning benchmarks such as MuSiQue, GSM8K, and AR-LSAT without the transformation. 
For short-answer QA, we first verify whether each model-generated prediction includes an answer in these benchmarks and then utilize GPT-4 to double-check whether the prediction is semantically correct. 
To generate predictions from the model, we utilize Prompt II in Section \ref{subsec:probing_set_construction}. 
The content of the GPT-4 prompt for prediction verification can be found in Table \ref{prompt:answer_verification}.

As shown in Table \ref{tab:general_reasoning}, models fine-tuned with NASA show enhanced or maintained general reasoning abilities while effectively tackling negative bias.
Since negative bias is a task-specific factor contributing to hallucination in binary decision tasks, its influence on general reasoning QA, like short-answer QA, is relatively low compared to other factors contributing to hallucination such as parametric knowledge. 
Therefore, improving reasoning ability in binary decision tasks does not necessarily enhance reasoning capabilities in general reasoning QA. 
This means that addressing bias in binary decision tasks while maintaining performance in general reasoning QA can be considered an advancement in the model's overall reasoning capability, not overfitted to the binary decision task.
% Model로부터 prediction을 생성하기 위해 우리는 앞서 언급된 Prompt II (Inquiring Parametric Knowledge)를 사용하였다. Prediction verification을 위한 GPT-4의 prompt 내용은 Table \ref{prompt:answer_verification}에서 확인할 수 있다.

%% file: sections/6.4_Enhanced_Model_Calibration.tex
\subsection{Enhanced Model Calibration}
% Figure \ref{fig:conf_hist}에서 확인하다시피, LLM의 negative bias를 이는 모델의 prediction에 대한 confidence와도 관련이 있다.
% Prediction confidence과 accuracy 사이의 calibration에 관한 연구는 LLM의 trustfulness를 강화하기 위해 필요하다.
% 우리는 NASA가 LLM의 yes-no question answering task에서의 calibration에 미치는 영향을 분석한다.

% Table \ref{tab:ece_results}는 Mistral-7B에 NASA를 적용했을 때 expected calibration error (\citealp{kadavath2022language}; \citealp{he2022preserving})를 보여준다.
% In this paper, 우리는 complex reasoning yes-no question answering dataset에서의 calibration에 집중했다.
% 다섯게 데이터셋에 대해서, NASA는 consistently 개선된 calibration을 보여준다. 
% 이는 NASA가 모델의 bias를 효과적으로 완화하여 prediction confidence와 accuracy가 더욱 align된 것으로 해석된다.
% \input{tables/calibration_test}
As Fig \ref{fig:conf_hist} illustrates, the negative bias of LLMs is also associated with the model's confidence in its predictions.
This issue regarding the gap between prediction confidence and actual accuracy is related to research on calibration, which aims to enhance the trustworthiness of LLMs (\citealp{kadavath2022language}). 
% We conduct an additional analysis of calibration errors when NASA is applied.
% 이러한 prediction confidence 와 실제 accuracy 간 gap에 관한 issue는 trustworthiness of LLMs를 enhance시키는 측면에 있어서 research on the calibration과 align되어 있다. 
% 따라서 우리는 NASA가 적용되었을 때 calibration error 측면에서 추가적인 분석을 수행한다.
% Consequently, research on the calibration between prediction confidence and accuracy is necessary to enhance the trustworthiness of LLMs (\citealp{kadavath2022language}).
Table \ref{tab:ece_results} shows that the expected calibration error (\citealp{he2022preserving}) is reduced when NASA is applied to Mistral.  
% Across five binary decision tasks, the NASA-tuned model demonstrates improved calibration. 
This indicates that NASA effectively mitigates model bias, better aligning prediction confidence and accuracy.

%% file: tables/general_reasoning.tex
\begin{table}[t!]
    \centering
    \caption{General QA reasoning ability of baseline (left of ``/'') and ours (right of ``/'').}
    {\resizebox{0.9\columnwidth}{!}{
    \renewcommand{\arraystretch}{1.1}
    \begin{tabular}{l|ccc}
    \toprule       
       \textbf{Model}   & \textbf{MuSiQue} & \textbf{GSM8K} & \textbf{AR-LSAT}\\ \hline
       \textbf{LLaMA} & 0.594 / 0.596 & 0.329 / 0.325 &  0.235 / 0.227 \\
       \textbf{Mistral} & 0.415 / 0.408 & 0.126 / 0.133 & 0.174 / 0.174  \\
       \textbf{Gemma} & 0.139 / 0.137 & 0.257 / 0.251 & 0.125 / 0.131  \\
       \textbf{Qwen} & 0.541 / 0.541 & 0.478 / 0.481 & 0.220 / 0.221 \\
     \bottomrule
    \end{tabular}
    }}    
    \label{tab:general_reasoning}
\end{table}

%% file: tables/calibration_test.tex
\begin{table}[t!]
    \centering
    \caption{Expected calibration error of baseline and ours.}
    {\resizebox{0.8\columnwidth}{!}{
    \renewcommand{\arraystretch}{1.1}
    \begin{tabular}{l|ccc}
    \toprule
      \textbf{Dataset}   & \textbf{Mistral} & \textbf{+ NASA} \\ \hline
       \textbf{StrategyQA} & 0.120 &  \textbf{0.116} \\
       \textbf{MuSiQue (Rephrased)} & 0.278 &  \textbf{0.212} \\
       % \textbf{2WikiMultiHopQA} & 0.212  & \textbf{0.066} \\
       \textbf{GSM8K (Rephrased)} & 0.301  &  \textbf{0.269} \\
       \textbf{MATH (Rephrased)} &  0.353 &  \textbf{0.260} \\
       \textbf{AR-LSAT (Rephrased)} & 0.399  & \textbf{0.355} \\
     \bottomrule
    \end{tabular}
    }}    
    \label{tab:ece_results}
\end{table}

% \begin{wraptable}{r}{0.5\textwidth}
%     \centering
%     \caption{Expected calibration error of baseline and ours.}
%     {\resizebox{0.5\textwidth}{!}{
%     \renewcommand{\arraystretch}{1.1}
%     \begin{tabular}{l|ccc}
%     \toprule
%       \textbf{Dataset}   & Mistral-7B & + NASA \\ \hline
%        \textbf{StrategyQA} & 0.120 &  \textbf{0.116} \\
%        \textbf{MuSiQue (Rephrased)} & 0.278 &  \textbf{0.212} \\
%        % \textbf{2WikiMultiHopQA} & 0.212  & \textbf{0.066} \\
%        \textbf{GSM8K (Rephrased)} & 0.301  &  \textbf{0.269} \\
%        \textbf{MATH (Rephrased)} &  0.353 &  \textbf{0.260} \\
%        \textbf{AR-LSAT (Rephrased)} & 0.399  & \textbf{0.355} \\
%      \bottomrule
%     \end{tabular}
%     }}    
%     \label{tab:ece_results}
% \end{wraptable}

%% file: sections/6.3_Analysis_on_few_shot_scenario.tex
\subsection{Analysis in a Few-shot Scenario}
We compare the F1 scores of the baselines and our models in a 4-shot setting on StrategyQA, rephrased GSM8K / AR-LSAT.
As shown in Table \ref{tab:few_shot}, we first observe that the F1 score gap between the baselines and our models decreases compared to the zero-shot scenario, as reported in Tables \ref{tab:main:multihop} and \ref{tab:main_math}.
Given that few-shot prompting has been reported to improve model calibration \cite{kadavath2022language}, these results suggest that the inclusion of few-shot examples enhances model calibration, thereby partially reducing negative bias.
Nevertheless, our model consistently outperforms the baselines in terms of F1 score across all cases.
This indicates that NASA remains effective in mitigating negative bias in few-shot scenarios.

%% file: sections/7_Discussion.tex
\section{Discussion} \label{7.conclusion}
Through our experiments, we identify two key characteristics of negative bias.

% \textbf{Negative bias is related to model calibration in binary decision tasks} 
% % As shown in Figure \ref{fig:conf_hist} and Tables \ref{tab:nas_nconfs_corr} and \ref{tab:few_shot}, 
% We demonstrate that negative bias is associated with the confidence of predictions.
% Table \ref{tab:ece_results} shows that NASA improves model calibration in binary decision tasks.
% These highlight the need for further research on negative bias from the perspective of model calibration.

\textbf{Model NAS may not negatively correlate with F1 score during fine-tuning} As shown in Table \ref{tab:appendix:q_only_results}, we observe instances where freezing certain parameters in the attention module yields higher F1 scores despite an increase in NAS.
These results suggest that a phase during fine-tuning exists where the negative correlation between the F1 score and NAS is disrupted.
To mitigate performance degradation, we introduce thresholds such as early stopping and halting within our framework.
\input{tables/few_shot}

\textbf{Accurate identification of negative attention heads is crucial, and NAS facilitates this process} Our ablation study in Table \ref{tab:appendix:random_head_results} indicates that fine-tuning randomly selected attention heads leads to suboptimal results.
In some cases, we observe abnormal increases or decreases in model NAS, implying the importance of fine-tuning problematic attention heads rather than unrelated parameters.
Leveraging NAS and the probing set, we successfully detect negative attention heads, leading to consistent improvements in the F1 score.

%% file: tables/few_shot.tex
\begin{table}[t!]
\centering
\caption{Precision, recall, and F1 scores in a 4-shot scenario (baseline / NASA).}
{\resizebox{\columnwidth}{!}{
\renewcommand{\arraystretch}{1.1}
\footnotesize
\begin{tabular}{@{}c|c|ccc}
\toprule
\textbf{Model} & \textbf{Dataset} & \textbf{Precision}  & \textbf{Recall}  & \textbf{F1}\\ 
\hline
\multirow{3}{*}{LLaMA} & StrategyQA & \textbf{0.899} / 0.872& 0.812 / \textbf{0.851}& 0.853 / \textbf{0.862}\\
& GSM8K  & \textbf{0.543} / 0.532& 0.720 / \textbf{0.870}& 0.619 / \textbf{0.661}\\
& AR-LSAT  & \textbf{0.562} / 0.557& 0.431 / \textbf{0.590}& 0.488 / \textbf{0.573}\\
\hline
\multirow{3}{*}{Mistral} & StrategyQA & \textbf{0.877} / 0.832& 0.824 / \textbf{0.889}& 0.850 / \textbf{0.859}\\
& GSM8K  & \textbf{0.548} / 0.522& 0.763 / \textbf{0.885}& 0.638 / \textbf{0.656}\\
& AR-LSAT  & \textbf{0.552} / 0.532& 0.589 / \textbf{0.720}& 0.570 / \textbf{0.612}\\
\hline
\multirow{3}{*}{Gemma} & StrategyQA & \textbf{0.903} / 0.897& 0.654 / \textbf{0.659}& 0.758 / \textbf{0.760}\\
& GSM8K  & 0.588 / \textbf{0.590}& 0.876 / \textbf{0.894}& 0.704 / \textbf{0.711}\\
& AR-LSAT  & \textbf{0.550} / 0.535& 0.366 / \textbf{0.405}& 0.439 / \textbf{0.461}\\
\hline
\multirow{3}{*}{Qwen} & StrategyQA & \textbf{0.901} / 0.894& 0.770 / \textbf{0.805}& 0.830 / \textbf{0.847}\\
& GSM8K  & \textbf{0.596} / 0.569& 0.765 / \textbf{0.827}& 0.670 / \textbf{0.674}\\
& AR-LSAT  & \textbf{0.535} / 0.546& 0.194 / \textbf{0.303}& 0.285 / \textbf{0.390}\\
\bottomrule
\end{tabular}
}}
\label{tab:few_shot}
\end{table}

%% file: sections/8_Conclusion.tex
\section{Conclusion} \label{8.conclusion}
% 우리는 large language model이 complex reasoning을 요구하는 binary decision task에서 치명적인 negative bias를 가진다는 문제를 발견한다. 해당 문제를 formulate하기 위한 negative attention score를 제안하고 이를 기반으로 query-agnostic한 negative head를 탐색하여 해당 head에 대한 parameter effienct tuning을 통해 효과적으로 bias 문제를 해결하는 NASA 방법론을 제안한다. 우리 방법론은 model의 성능 향상 뿐만 아니라 interpretability의 측면에서도 유용한 분석 framework로서 작용할 수 있을 것이다.

We identify a critical issue where large language models exhibit negative bias in binary decision tasks requiring complex reasoning. To address this issue, we propose a negative attention score and employ it to discover query-agnostic negative heads. By performing parameter-efficient tuning on these heads, we introduce the NASA method, which effectively mitigates the bias problem. Our method not only enhances the performance of the model but also serves as a useful analytical framework from the perspective of interpretability.

%% file: sections/Limitations.tex
\section*{Limitations}
In this work, we focus on understanding the relationship between negative biases exhibited in fine-tuned LLMs and attention heads. 
However, further research is still needed to comprehend the causes and mechanisms behind the occurrence of negative biases in LLMs, and we anticipate that our observations and experimental results will lay the groundwork for future work.
Additionally, we have adopted a scheme of fine-tuning a small number of attention heads individually. 
In future work, it may be possible to explore more time-efficient training methods.
While this study focuses on understanding the characteristics of negative bias attention heads, future work could involve a more integrated research approach connecting various elements.

%% file: sections/A_data_Statistics.tex
\section{Data Statistics} \label{appen:data_statistics}
\input{tables/data_statistics}
Table \ref{tab:data_statistics} shows the statistics of dataset utilized in our work.

%% file: tables/data_statistics.tex
\begin{table}[t!]
    \centering
    \caption{Data statistics of datasets utilized in our work.}
    {\resizebox{\columnwidth}{!}{
    \renewcommand{\arraystretch}{1.1}
    \begin{tabular}{l|ccc}
    \toprule
      \textbf{Data}   & \# of Positives & \# of Negatives & \# of total\\ \hline
       \textbf{StrategyQA} & 1071 &  1219 & 2290 \\
       \textbf{MuSiQue (Rephrased)} & 1208 &  1208 & 2416 \\
       \textbf{GSM8K (Rephrased)} & 1001  &  999 & 2000 \\
       \textbf{MATH (Rephrased)} &  1000 &  1000 & 2000 \\
       \textbf{AR-LSAT (Rephrased)} & 820  & 777 & 1597 \\
     \bottomrule
    \end{tabular}
    }}   
    \label{tab:data_statistics}
\end{table}

%% file: sections/A_statistical_analysis.tex
\section{A Statistical Analysis of the Relationship between Negative Confidence and NAS} \label{appen:statistical_analysis}
\input{tables/statistical_analysis}
To analyze the impact of negative confidence on model NAS, we conduct a multiple regression analysis by including positive confidence as a variable, as shown in Table \ref{tab:statistical_analysis}.
In all cases except for LLaMA, where the corrected $p$-value is high, negative confidence is observed to have a statistically significant positive correlation with model NAS.
Meanwhile, considering the effect sizes, future work could investigate additional factors influencing model NAS.

%% file: tables/statistical_analysis.tex
\begin{table}[t!]
    \centering
    \caption{Results of multiple regression alongside a correlation for multiple comparisons. $r^2$ denotes the effect size.}
    {\resizebox{0.9\columnwidth}{!}{
    \renewcommand{\arraystretch}{1.1}
    \begin{tabular}{l|cccc}
    \toprule
      \textbf{Model}   & \textbf{Coefficient} & \textbf{Corrected $p$-value} & \textbf{$r^2$}\\ \hline
       \textbf{LLaMA} & $-$3.04 &  7.18$E-01$ & 0.271 \\
       \textbf{Mistral} & 19.57 &  1.22$E-03$ & 0.060 \\
       \textbf{Gemma} & 29.61  &  1.23$E-21$ & 0.377 \\
       \textbf{Qwen} &  126.03 &  4.65$E-69$ & 0.265 \\
     \bottomrule
    \end{tabular}
    }}   
    \label{tab:statistical_analysis}
\end{table}

%% file: sections/C_Instructions.tex
\section{Instructions for NASA} \label{C_instructions}
Table \ref{prompt_instructions} presents the instructions utilized for NASA process in our work and Table \ref{tab:transferability}.

%% file: sections/details_of_binary_decision_data_construction.tex
\section{Details of Binary Decision Data Construction}\label{sec:appendix:details_of_binary_decision_data_construction}
To convert general QA datasets into binary decision tasks, we create positive and negative examples where the answers are ``Yes'' and ``No'', respectively. 
Note that StrategyQA is a yes-no QA dataset and does not require a transformation. 

MuSiQue, GSM8K, and MATH are short-answer QA datasets, and AR-LSAT is a multiple-choice dataset. For short-answer QA datasets, we use Prompt I in Section \ref{subsubsec:binary_decision} to generate positive examples. 
AR-LSAT, a multiple-choice dataset, utilizes the positive question generation prompt from Table \ref{prompt:appendix:wrong_label_generation} to create questions with the answer "Yes." Consequently, StrategyQA and AR-LSAT (Rephrased) are yes-no QA datasets, while MuSiQue (Rephrased), GSM8K (Rephrased), and MATH (Rephrased) is the form of answer verification.

\subsection{GPT-4 Prompts for Negative Example Construction} 
To construct negative examples, we utilize GPT-4. 
In short-answer QA (i.e., MuSiQue, GSM8K, and MATH), we use the wrong label generation prompt from Table \ref{prompt:appendix:wrong_label_generation} to obtain incorrect answers. 
We then replace the label in Prompt I with the incorrect answer to generate a negative example.

In multiple choice (i.e., AR-LSAT), we use the negative question generation prompt from Table \ref{prompt:appendix:wrong_label_generation} to create questions with the answer ``No''. 
An example of a rephrased AR-LSAT example can be found in Table \ref{tab:appendix:example_of_datasets}.

%% file: sections/details_of_parametric_sample_selection.tex
\section{Details of Parametric Sample Selection}\label{sec:appendix:details_of_parametric_sample_selection}
As mentioned in Section \ref{subsec:probing_set_construction}, we construct the probing and fine-tuning sets using the HotpotQA dataset, which is a short-answer QA dataset.
From the HotpotQA dataset, we use samples that can be correctly answered solely with parametric knowledge, without any context, as our probing samples. Specifically, we suppose that the probing samples must meet the condition: \textit{the question should be answerable using the model’s parametric knowledge.}
To determine whether the question in the sample meets the condition, we construct the prompt as follows:
\begin{tcolorbox}[
    enhanced, 
    arc=0pt, 
    outer arc=0pt, 
    boxsep=0pt, 
    left=0pt, 
    right=0pt, 
    top=0pt, 
    bottom=0pt, 
    colback=darkgray, 
    colframe=darkgray
]
  % Upper sub-box with dark gray background and shorter height
  \begin{tcolorbox}[
      colback=darkgray, 
      colframe=darkgray, 
      coltext=white, 
      arc=0pt, 
      outer arc=0pt, 
      boxsep=5pt, 
      left=5pt, 
      right=5pt, 
      top=5pt, 
      bottom=-0.5pt, 
      height=0.35cm, % Set the desired height for the upper box
      valign=center, % Vertically center the content
      borderline south={0pt}{0pt}{darkgray}
  ]
    Prompt II (Inquiring Parametric Knowledge)
  \end{tcolorbox}
  % Lower sub-box with light gray background and vertically centered text
  \begin{tcolorbox}[
      colback=lightgray, 
      colframe=lightgray, 
      coltext=black, 
      arc=0pt, 
      outer arc=0pt, 
      boxsep=5pt, 
      left=5pt, 
      right=5pt, 
      top=1pt, 
      bottom=1pt, 
      height=1.5cm,
      valign=center, % Vertically center the content
      borderline north={0pt}{0pt}{lightgray}
  ]
  % \footnotesize
    % \texttt{{\char`\\}c} Q: {\char`\\}? Options: {\char`\\}o A:
    You MUST answer shortly the given question based on your knowledge.
    Question: $\{question\}$
    Answer: 
  \end{tcolorbox}
\end{tcolorbox}
% 이 prompt는 모델의 negative bias의 영향을 최소화하면서 parametric knowledge에 속하는 sample을 효과적으로 선별할 수 있게 한다.
This prompt is designed to identify attention heads that attend to the negative token in the instruction even when a positive answer is expected. 
We assume that if attention heads attend to the negative token when the model knows the correct answer and the target answer is ``Yes'', it is generated by a negative bias in the model.
% Condition II를 따지기 위한 prompt II는 다음과 같이 구성한다.
This prompt effectively selects samples that belong to the model’s parametric knowledge while minimizing the influence of the model's negative bias.
We assume that samples meet the condition if the model returns the correct answers for the prompt, and we use them as the probing set.
We utilize GPT-4 to verify whether the prediction matches the label. 

During the probing sample selection process, we first input the question into the target model to extract a prediction. 
Since the HotpotQA dataset can contain various forms of answers, we use GPT-4 to determine if the prediction corresponds to the label. 
Specifically, we input the prompt from Table \ref{prompt:answer_verification} into GPT-4 and only include samples in the probing set if the response is ``Yes''.

%% file: sections/NASA_Algorithm.tex
\section{Algorithm of Incremental Head Tuning}\label{appendix:nasa_algorithm}
Algorithm \ref{alg:nasa_tuning} shows the details of head-wise incremental tuning of NASA.

%% file: sections/neg_response_analysis.tex
\section{Case Study for Negative Responses} \label{appendix:neg_res_analysis}

In this section, we classify negative responses based on the following criteria:

Firstly, in a general QA reasoning task, a model's response to a specific question can be broadly categorized into below two cases.

\begin{itemize}
    \item Deterministic Case (Det): The model provides a certain answer regardless of its correctness.    
    \item Non-Deterministic Case (Non-Det): The model fails to provide a certain answer, indicating that the question is unanswerable due to insufficient information from the given context or its knowledge base.
\end{itemize}

Also, the deterministic case can be further classified based on whether the certain answer provided by the model is correct (True) or incorrect (False). Considering this, a model's response to a general question requiring a short answer can be categorized into the following three types.

\begin{itemize}
    \item True Deterministic (Det-T): The model provides a correct answer.  
    \item False Deterministic (Det-F): The model provides an incorrect answer.
    \item Non-Deterministic (Non-Det): The model indicates that the question is unanswerable.
\end{itemize}

\input{algorithms/nasa_tuning}

Additionally, beyond the extrinsic responses of the model, we can classify the model's responses based on the intrinsic factor of confidence. Specifically, if the model exhibits low confidence in the response, it can be considered closer to not knowing the answer to the question. We calculate the entropy of the output distribution for the first token of the model's response and use the median value of these entropies across all responses to classify them as high-confidence (low-entropy) or low-confidence (high-entropy). This classification is performed based on the original model's responses before applying NASA method.

We apply this classification to cases where the model gives a negative response in a binary decision task. Negative responses can be classified as true negative (TN) or false negative (FN), with our focus in this discussion on FN. In this, an FN response classified as high-confident Det-T is the case where the model knows the answer is correct but says ``No'', while an FN response classified as Non-Det is the case where the model doesn't know the correct answer and says ``No''. For high-confident Det-F responses, the model does not indicate unanswerability but rather holds incorrect knowledge. This issue falls outside the scope of our methodology, which focuses on handling negative bias in binary decision tasks rather than general knowledge-intensive tasks.

Although it is clear that the case where the model knows the answer is correct but says ``No'' is the most problematic, we argue that in a binary decision task, the case where the model doesn't know the correct answer and says ``No'' can be related to the issue of negative bias. In a binary decision task where the model must answer ``Yes'' or ``No'', instances where the model is uncertain about the correct answer should result in an equal frequency of positive (yes) and negative (no) responses. However, our analysis indicates that existing models tend to output negative responses more confidently and frequently (as shown in Figures 1 and 2 in the paper). This suggests that when the model is uncertain, it predominantly outputs negative responses, which negatively impacts the trustworthiness of these negative responses. Therefore, for models exhibiting such negative bias, Non-Det and low-confident responses should also be partially adjusted.

Table \ref{tab:fn_composition} shows the compositions of FN responses from the original models according to the classifications mentioned. 
% It is evident that in all types of reasoning tasks addressed in our work (multi-hop, mathematical, logical reasoning), the FN responses include both cases: (A) the model knows the answer is correct but says “no” and (B) the model doesn’t know the correct answer and says “no”. 
It is evident that across all types of reasoning tasks examined in our work, including multi-hop, mathematical, and logical reasoning, FN responses encompass two distinct cases: (A) instances where the model recognizes the correct answer but incorrectly responds with ``No'', and (B) instances where the model lacks knowledge of the correct answer and responds with ``No''.

Meanwhile, for Gemma, we observe that the model tends to generate overly cautious responses when performing multi-hop QA. Specifically, instead of selecting the ``unanswerable'' option for multiple samples, Gemma rephrases its responses in an alternative long-form manner. As a result, instances that should have originally been classified as Non-Det are instead categorized as Det-F. This should be taken into account when interpreting the results in Table \ref{tab:fn_composition}.
% 한편, for Gemma, 우리는 model이 전반적인 multi-hop QA를 수행할 때 지나치게 조심스러운 응답을 출력하는 현상을 발견했다. 구체적으로, Gemma는 다수의 samples에 대해서 unanswerable option을 선택하는 대신 다른 방식으로 풀어서 응답하는 현상을 발견했다. 이에 따라 원래 Non-Det case에 해당되어야 할 이들이 Det-F로 분류되었음을 참고하면 좋겠다.

% Specifically, for Gemma, the model often outputs responses indicating that many questions in multi-hop QA are unanswerable. Please note that these are classified as Det-F in the table.

%% file: algorithms/nasa_tuning.tex
\begin{algorithm}[t!]
\caption{Head-wise Incremental Tuning}\label{alg:nasa_tuning}
\textbf{Input: } {Fine-tuning set $D$, validation set $V$, model $\psi$, single head NAS threshold $\rho$, list of negative attention heads to be fine-tuned $P$, early halting threshold $\tau$, maximum training epoch $T$, set of all layer and head indices $L$ and $H$, respectively.}\\
\textbf{Output: } {NASA-tuned model.} \\
\begin{algorithmic}[1]
\newcommand*{\Break}{\textbf{break}}
% \State Let $P = \{(l_1, h_1), (l_2, h_2), \dots, (l_N, h_N)\}$ represent the list of layer and head indices of attention heads sorted in descending order by single head NAS on probing set.
\For{each $(l, h) \in P$}
    \State $\psi_{init} = \psi$
    % \State $\alpha_{init} = \frac{1}{|V|} \sum_{v_j \in V} \text{NAS}^{l_i,h_i}_{v_j}$
    \State $\alpha_{init} = \text{NAS}(V, l, h)$ % \Comment{Initial single head NAS}
    % \State $\beta_{init} = \sum\sum_{l \in L, h \in H}\frac{1}{|V|} \sum_{v_j \in V} \text{NAS}^{l,h}_{v_j}$
    \State $\beta_{init} = \text{NAS}(V, L, H)$ % \Comment{Initial model NAS}
    \State $\alpha=\beta=\infty$
    \For{epoch $= 1$ to $T$}
        \State Update $\psi$ by fine-tuning $h$-th \\ attention head in the $l$-th layer using $D$.
        % \State $\alpha' = \frac{1}{|V|} \sum_{v_j \in V} \text{NAS}^{l_i,h_i}_{v_j}$
        \State $\alpha' = \text{NAS}(V, l, h)$ % \Comment{Current single head NAS}
        % \State $\beta' = \sum\sum_{l \in L, h \in H}\frac{1}{|V|} \sum_{v_j \in V}
        \State $\beta' = \text{NAS}(V, L, H)$ % \Comment{Current model NAS}
        \If{$\alpha' > \alpha$ \textbf{or} $\alpha' < \rho$ \textbf{or} $\beta' > \beta$}
            \State \textbf{break} \Comment{Early stopping}
        \EndIf
        \State $\alpha \gets \alpha', \beta \gets \beta'$
    \EndFor
    \If{$\alpha' > \alpha_{init}$ or $\beta' > \beta_{init}$}
        \State $\psi \gets \psi_{init}$ \Comment{Update cancellation}
    \EndIf
    \If{$\beta' < \tau$}
        \State \textbf{break} \Comment{Early halting}
    \EndIf
    
\EndFor
\State \Return $\psi$
\end{algorithmic}
\end{algorithm}

%% file: sections/response_shift_ratio_analysis.tex
\section{Analysis of Negative Responses Shifting in NASA} \label{appendix:res_shift_ratio_analysis}

In this section, we analyze the ratios of samples which are originally responded to as negative by the original model and subsequently changed to positive by the NASA model.  We categorize the ratio cases by each classification group explained in section \ref{appendix:neg_res_analysis}.

Table \ref{tab:fn_to_tp} illustrates the ratios of samples shifted from FN to TP. This demonstrates that the Det-T corresponding to the case where the model knows the answer is correct but says ``No'' generally has the highest rate of change. This finding indicates that our methodology does not indiscriminately add positive bias to the model but rather enhances the model's binary decision reasoning ability effectively.

Table \ref{tab:tn_to_fp} illustrates the ratios of samples shifted from TN to FP. As discussed in the section on main results, some samples that are incorrectly categorized as TN due to negative bias might shift to FP (Det-F), and this case does not correspond to the model degradation. Actually, among the TN samples, those classified as Det-T with high confidence exhibit a lower shift ratio compared to their Det-F counterparts. This indicates that the model tends to maintain predictions for samples that it can accurately classify. In conclusion, it can be interpreted that NASA has improved reasoning capability in binary decision tasks while maintaining general reasoning capability.

%% file: sections/6.2_Generalization_to_Universal_Binary_Decision.tex
\section{Generalization to Universal Binary Decision}\label{appendix:generalization_to_universal_binary_decision}
We analyze the generalization ability for binary decisions by replacing the positive and negative vocabulary used in our fine-tuning set.
Our fine-tuning set consists of ``Yes'' and ``No'' in the instruction, and we test with the tokens ``True'' and ``False'', as well as ``Correct'' and ``Wrong''.
\input{tables/universal_binary_decision}
Table \ref{tab:universal_binary_decision} presents the results of experiments with LLaMA on three datasets transformed for the binary decision task.
We observe that models fine-tuned with NASA achieve semantic-level generalization without overfitting to the specific vocabulary.

%% file: tables/universal_binary_decision.tex
\begin{table}[t!]
\centering
\caption{Accuracy and F1 (left and right of ``/'', respectively) scores on various positive and negative tokens.}
{\resizebox{0.9\columnwidth}{!}{
\renewcommand{\arraystretch}{1.1}
\footnotesize
\begin{tabular}{@{}c|c|cc}
\toprule
\textbf{Type} & \textbf{Dataset} & \textbf{LLaMA}  & \textbf{+ NASA} \\ 
\hline
\multirow{3}{*}{\parbox{1.3cm}{True/False}} & StrategyQA  & 0.855 / 0.828  & \textbf{0.866} / \textbf{0.853} \\
& MATH  & 0.559 / 0.452  & \textbf{0.574} / \textbf{0.583} \\
& AR-LSAT  & \textbf{0.512} / 0.306  & 0.506 / \textbf{0.405} \\
\hline
\multirow{3}{*}{\parbox{1.7cm}{Correct/Wrong}} & StrategyQA  & \textbf{0.852} / 0.838  & 0.846 / \textbf{0.840} \\
& MATH  & \textbf{0.545} / 0.660  & 0.544 / \textbf{0.675} \\
& AR-LSAT  & \textbf{0.523} / 0.540  & 0.522 / \textbf{0.590} \\
\bottomrule
\end{tabular}
}}
\label{tab:universal_binary_decision}
\end{table}

%% file: sections/6.3_Transferability_across_Various_Instructions.tex
\section{Transferability across Various Instructions}\label{appendix:transferrability_across_various_instructions}
\input{tables/transferability}
% In this work, 우리는 방법론을 적용하는 전체 process에서 binary decision: Yes-No QA를 위한 instruction을 고정한다. In this section, 우리는 해당 instruction이 아닌 paraphrase되어 다른 형태의 content를 담고 있는 binary decision instruction이 inference에서 주어지는 경우에서도 우리 method가 robust한지 평가한다. 우리는 NASA 방법론 적용 시에 사용한 instruction을 paraphrase한 instruction type A,B에 대해서 LLaMA3-8B과 NASA를 적용한 경우의 MuSiQue, GSM8K, AR-LSAT benchmark에서의 binary decision 성능을 평가한다. Table \ref{tab:transferability}를 통해서 prompt A, B 모두에서 NASA가 적용된 방법론이 여전히 baseline보다 더 향상된 성능을 보임을 확인할 수 있다. 이는 우리 방법론의 instruction robustness를 주장한다.

We fix the instruction for binary decision-making throughout the application of our method. 
In this section, we evaluate the robustness of our method when presented with instructions in the inference stage that are different forms of content. 
We assess the binary decision performance using the LLaMA with and without NASA on benchmarks: MuSiQue, GSM8K, and AR-LSAT using paraphrased instruction types A and B. 
Note that the datasets are transformed for the binary decision task.
Details of instruction types can be found in Appendix \ref{C_instructions}.
As shown in Table \ref{tab:transferability}, models fine-tuned with NASA still demonstrate superior performance compared to the baseline in both prompt types. 
This supports our claim of instruction robustness for our method. 
Fig \ref{tab:transferability} in Appendix \ref{C_instructions} shows the content of instructions.

%% file: tables/transferability.tex
\begin{table}[t!]
\centering
\caption{Accuracy and F1 (left and right of ``/'', respectively) scores on various instruction types.}
{\resizebox{0.9\columnwidth}{!}{
\renewcommand{\arraystretch}{1.1}
\footnotesize
\begin{tabular}{@{}c|c|cc}
\toprule
\textbf{It. Type} & \textbf{Dataset} & \textbf{LLaMA}  & \textbf{+ NASA} \\ 
\hline
\multirow{3}{*}{A} & MuSiQue & 0.686/0.641& \textbf{0.710}/\textbf{0.700}\\
& GSM8K  & 0.534/0.472& \textbf{0.543}/\textbf{0.575}\\
& AR-LSAT  & \textbf{0.508}/0.306& 0.507/\textbf{0.370}\\
\hline
\multirow{3}{*}{B} &  MuSiQue  & 0.708/0.692& \textbf{0.736}/\textbf{0.756}\\
& GSM8K  & \textbf{0.536}/0.611& 0.531/\textbf{0.665}\\
& AR-LSAT  & 0.512/0.382& \textbf{0.518}/\textbf{0.500}\\
\bottomrule
\end{tabular}
}}
\label{tab:transferability}
\end{table}

%% file: sections/Ablation_Study.tex
\section{Ablation Study} \label{appen:ablation_study}

NASA updates the parameters related to the inference of attention weights of negative attention heads (i.e., query and key projection weights).
To demonstrate the effectiveness of NASA, we conduct two ablation studies: i) updating only the query projection weight and ii) random attention head tuning.

\subsection{Freezing Key Projection Weight}
We perform fine-tuning using the same pipeline as before, except that we freeze the key projection weight. 
Note that strategies such as early stopping and halting, as well as other hyperparameters, remain the same as those used in the NASA settings.

The experimental results of NASA and freezing key projection weights are shown in Table \ref{tab:appendix:q_only_results}. 
In general, both the F1 score and NAS are relatively better for NASA, while the trend in accuracy is not distinctly observable. 
An exceptional case is Qwen at AR-LSAT, where NASA’s accuracy and F1 score are significantly lower than those of its counterpart.

\subsection{Random Attention Head Tuning}
In another ablation study, we experiment by randomly selecting attention heads to be fine-tuned.
As shown in Table \ref{tab:appendix:random_head_results}, NAS typically outperforms most others in terms of F1 score and expected calibration error.
Notably, during random attention head tuning, model NAS largely diverges in the case of LLaMA and Mistral.
These results imply that carefully selecting attention heads is critical and our framework effectively detects and addresses negatively biased attention heads.

%% file: sections/confidence_histogram_analysis.tex
\section{Confidence Histogram Analysis}\label{sec:appendix:confidence_histogram_analysis}
To analyze the confidence of binary decisions between the baseline and our model, we plot confidence histograms for the LLaMA model, similar to Fig \ref{fig:conf_hist}.
In Fig \ref{fig:conf_hists_llama}, we observe that NASA influences the prediction confidence of LLMs in binary decision tasks.
In the baseline model, except for StrategyQA, we frequently observe the return of negative responses with high confidence.
On the other hand, after incorporating NASA, the model demonstrates a reduction in the frequency and confidence gap between positive and negative responses.

Meanwhile, for datasets of mathematical reasoning, there has been a significant increase in the frequency of positive responses, even though F1 scores also increased.
As discussed in Section \ref{subsec:main_results}, future research could explore additional regularization during the fine-tuning process.

%% file: sections/D_Samples.tex
\section{Samples of Attention Heads} \label{samples}
Fig \ref{fig:attn_head_llama} presents the real examples of attention weights of heads in the LLaMA model for a GSM8K rephrased sample. Fig \ref{fig:attn_head_nasa} presents the real examples of attention weights of heads in the LLaMA model with NASA for the same GSM8K rephrased sample. 

%% file: sections/Computation.tex
\section{Computation}
For all fine-tuning experiments, we use 2 NVIDIA A40 GPUs for approximately 3 hours.

%% file: sections/License.tex
\section{Licenses}
StrategyQA, MuSiQue, MetaMATH, and AR-LSAT are under the license of MIT license, CC-BY-4.0 license, MIT License, and MIT license, respectively.
LLaMA3-8B-Instruct, Mistral-7B-Instruct-v0.3, Gemma-1.1-7b-it, and Qwen2-7B-Instruct, and GPT-4 are under the license of META LLAMA 3 COMMUNITY LICENSE AGREEMENT, Apache License 2.0, Gemma, Apache License 2.0, and OpenAI, respectively.

%% file: sections/E_AI_Assistant.tex
\section{Usage of AI Writing Assistance} \label{ai_Writing}

This paper received linguistic assistance from the AI assistant GPT-4, which provided services including paraphrasing, spell-checking, and refinement of the original content authored. No additional help was utilized beyond this support.

\input{prompts/instructions}
\input{prompts/wrong_label_generation}
\input{tables/example_of_datasets}
\input{prompts/parametric_answer_verification}

\input{tables/fn_compositions}
\input{tables/fn_to_tp}
\input{tables/tn_to_fp}

\input{tables/q_only_results}
\input{tables/random_head_tuning}

\input{figures/conf_hists_others_}

\input{figures/attn_heads_llama_}
\input{figures/attn_heads_nasa_}

%% file: prompts/instructions.tex
\begin{table*}
\centering
\caption{Instructions used NASA process in our work and Table \ref{tab:transferability}.}
{\small
\renewcommand{\arraystretch}{1.2}
\begin{tabular}{p{15cm}}
\toprule
\textbf{Instruction in NASA}

\texttt{You are given a question and you MUST answer Yes or No.}
\\ \hline
\textbf{Instruction type A}

\texttt{You are asked a question that demands a clear Yes or No answer.}
\\ \hline
\textbf{Instruction type B}

\texttt{A question is posed to you, and you are obligated to answer either Yes or No.}
\\ \bottomrule
\end{tabular}
}
\label{prompt_instructions}
\end{table*}

%% file: prompts/wrong_label_generation.tex
\begin{table*}
\centering
\caption{GPT-4 prompts used in wrong label or question generation for binary decision sample construction.}
{\small
\renewcommand{\arraystretch}{1.2}
\begin{tabular}{p{15cm}}
\toprule
\textbf{Wrong Label Generation (MuSiQue, GSM8K, and MATH)}

\texttt{[System]} \\
\texttt{Generate the short wrong answer word for the given question. You MUST refer the given context about the question.} \\
\texttt{[User]} \\
\texttt{Question:} \{\textit{question}\} \\
\texttt{Context:} \{\textit{context}\} \\
\texttt{Answer:} \{\textit{label}\} \\
\texttt{Wrong answer:}
\\ \hline
\textbf{Positive / Negative Question Generation (AR-LSAT)}

\texttt{[System]} \\
\texttt{Convert the given question to the binary Yes-No question based on the context about the given question and the answer of the given question. The answer of the converted question must be YES. Do NOT omit the condition in the given question like `If ...' or `Suppose ...'. You MUST include the entire contents of the given question to the converted question.} \\
\texttt{[User]} \\
\texttt{Context:} \{\textit{context}\} \\
\texttt{Question:} \{\textit{question}\} \\
\texttt{Answer:} \{\textit{label / wrong label}\} \\
\texttt{Converted Question: }
\\ \bottomrule
\end{tabular}
}
\label{prompt:appendix:wrong_label_generation}
\end{table*}

%% file: tables/example_of_datasets.tex
\begin{table*}
\centering
\caption{Examples of original datasets and rephrased AR-LSAT. Examples are sampled from StrategyQA (\citealp{geva2021strategyqa}), MuSiQue (\citealp{trivedi2022musique}), GSM8K-Rephrased and MATH-Rephrased (\citealp{yu2023metamath}), and AR-LSAT (\citealp{zhong2021arlsat}).}
{\small
\renewcommand{\arraystretch}{1.2}
\begin{tabular}{|p{15cm}|}
\hline
\textit{StrategyQA} \\
\hline 
\textbf{Context}: Depression is caused by low levels of serotonin, dopamine and norepinephrine. Monoamine Oxidase breaks down neurotransmitters and lowers levels of serotonin, dopamine and norepinephrine.\\
\textbf{Question}: Would a Monoamine Oxidase candy bar cheer up a depressed friend? \\
\textbf{Answer}: No 
\\ \hline
\textit{MuSiQue (Original)} \\
\hline 
\textbf{Context}: [Title: President of the Confederate States of America] The president was indirectly elected by the people through the Electoral College to a six - year term and was one of only two nationally elected Confederate officers ... \\
\textbf{Question}: When did the president of the Confederate States of America end his fight in the Mexican-American war? \\
\textbf{Answer}: 1848
\\ \hline
\textit{GSM8K (Original)} \\
\hline 
\textbf{Question}: What is the total cost of purchasing equipment for all sixteen players on the football team, considering that each player requires a \$25 jersey, a \$15.20 pair of shorts, and a pair of socks priced at \$6.80? \\
\textbf{Answer}: 1500
\\ \hline
\textit{MATH (Original)} \\
\hline 
\textbf{Question}: What is the sum of all positive integer values of $n$ for which $\frac{n+6}{n}$ is an integer? \\
\textbf{Answer}: 23
\\ \hline
\textit{AR-LSAT (Original)} \\
\hline 
\textbf{Context}: Hannah spends 14 days, exclusive of travel time, in a total of six cities. Each city she visits is in one of three countries—X, Y, or Z. Each of the three countries has many cities. Hannah visits at least one city in each of the three countries. She spends at least two days in each city she visits. She spends only whole days in any city. If the city of Nomo is in country X, and if Hannah spends as many days as possible in Nomo and as few days as possible in each of the other cities that she visits \\
\textbf{Question}: Which one of the following must be true? \\
(A) Hannah cannot visit any other cities in country X. \\
(B) Hannah can visit four cities in country Y. \\
(C) Hannah can spend six days in Nomo. \\
(D) Hannah cannot spend more than four days in country Z. \\
(E) Hannah can visit, at most, a total of four cities in countries Y and Z. \\
\textbf{Answer}: (B)  Hannah can visit four cities in country Y.
\\ \hline
\textit{AR-LSAT (Rephrased)} \\
\hline 
\textbf{Question}: Hannah spends 14 days, exclusive of travel time, in a total of six cities. Each city she visits is in one of three countries—X, Y, or Z. Each of the three countries has many cities. Hannah visits at least one city in each of the three countries. She spends at least two days in each city she visits. She spends only whole days in any city. If the city of Nomo is in country X, and if Hannah spends as many days as possible in Nomo and as few days as possible in each of the other cities that she visits can Hannah visit four cities in country Y? \\
\textbf{Answer}: Yes \\
\hline
\end{tabular}
}
\label{tab:appendix:example_of_datasets}
\end{table*}

%% file: prompts/parametric_answer_verification.tex
\begin{table*}
\centering
\caption{GPT-4 prompt used in prediction verification for the general reasoning task.}
{\small
\renewcommand{\arraystretch}{1.2}
\begin{tabular}{p{15cm}}
\toprule
\texttt{[System]} \\
\texttt{Can the prediction be considered as the same meaning as the answer to the question? You must answer only yes or no.} \\
\texttt{[User]} \\
\texttt{Question:} \{\textit{context}\} \\
\texttt{Prediction:} \{\textit{model prediction}\} \\ 
\texttt{Answer:} \{\textit{label}\}
\\ \bottomrule
\end{tabular}
}
% during the selection of parametric samples from HotpotQA.}
\label{prompt:answer_verification}
\end{table*}

%% file: tables/fn_compositions.tex
\begin{table*}[!t]
\centering
\caption{Composition of false negative samples of original models in the rephrased datasets. High denotes high-confident and Low denotes low-confident.}
% {\resizebox{\textwidth}{!}{
\begin{tabular}{l|c||cccccc}
\toprule
\multirow{2}{*}{\textbf{Dataset}} & \multirow{2}{*}{\textbf{Model}} & \multicolumn{2}{c}{\textbf{Det-T}}                                   & \multicolumn{2}{c}{\textbf{Det-F}}                                   & \multicolumn{2}{c}{\textbf{Non-Det}}                                 \\
                                  &                               & \textbf{High} & \textbf{Low} & \textbf{High} & \textbf{Low} & \textbf{High} & \textbf{Low} \\ \hline
\multirow{4}{*}{MuSiQue} & LLaMA                                   & 0.21                              & 0.182                            & 0.141                             & 0.261                            & 0.04                              & 0.166                            \\
                                  & Mistral                                  & 0.159                             & 0.103                            & 0.218                             & 0.27                             & 0.095                             & 0.155                            \\
                                  & Gemma                                    & 0.009                             & 0.022                            & 0.546                             & 0.42                             & 0                                 & 0.003                            \\
                                  & Qwen                                    & 0.25                              & 0.174                            & 0.139                             & 0.327                            & 0.007                             & 0.104                            \\ \hline
\multirow{4}{*}{GSM8K}   & LLaMA                                  & 0.209                             & 0.092                            & 0.286                             & 0.414                            & 0                                 & 0                                \\
                                  & Mistral                                  & 0.076                             & 0.03                             & 0.407                             & 0.486                            & 0.001                             & 0                                \\
                                  & Gemma                                    & 0.097                             & 0.065                            & 0.378                             & 0.459                            & 0                                 & 0                                \\
                                  & Qwen                                    & 0.194                             & 0.121                            & 0.278                             & 0.407                            & 0                                 & 0                                \\ \hline
\multirow{4}{*}{AR-LSAT} & LLaMA                                   & 0.057                             & 0.06                             & 0.303                             & 0.312                            & 0.158                             & 0.11                             \\
                                  & Mistral                                  & 0.058                             & 0.078                            & 0.385                             & 0.437                            & 0.002                             & 0.04                             \\
                                  & Gemma                                    & 0.011                             & 0.025                            & 0.49                              & 0.474                            & 0                                 & 0                                \\
                                  & Qwen                                    & 0.068                             & 0.108                            & 0.306                             & 0.413                            & 0.092                             & 0.013                      \\     
\bottomrule
\end{tabular}
% }}
\label{tab:fn_composition}
\end{table*}

%% file: tables/fn_to_tp.tex
\begin{table*}[!t]
\centering
\caption{Response shift ratio by NASA on false negative samples (FN $\rightarrow$ TP ratio) in the rephrased datasets. High denotes high-confident and Low denotes low-confident.}
% {\resizebox{\textwidth}{!}{
\begin{tabular}{l|c||cccccc}
\toprule
\multirow{2}{*}{\textbf{Dataset}} & \multirow{2}{*}{\textbf{Model}} & \multicolumn{2}{c}{\textbf{Det-T}}                                   & \multicolumn{2}{c}{\textbf{Det-F}}                                   & \multicolumn{2}{c}{\textbf{Non-Det}}                                 \\
                                  &                               & \textbf{High} & \textbf{Low} & \textbf{High} & \textbf{Low} & \textbf{High} & \textbf{Low} \\ \hline
\multirow{4}{*}{MuSiQue} & LLaMA                                   & 0.364& 0.39& 0.268& 0.326& 0.136& 0.197
\\
                                  & Mistral                                  & 0.432& 0.449& 0.265& 0.28& 0.203& 0.163
\\
                                  & Gemma                                    & 0.185& 0& 0.029& 0& 0& 0
\\
                                  & Qwen                                    & 0.122& 0.15& 0.074& 0.093& 0& 0.032
\\ \hline
\multirow{4}{*}{GSM8K}   & LLaMA                                  & 0.662& 0.694& 0.629& 0.715& N/A& N/A
\\
                                  & Mistral                                  & 0.817& 0.667& 0.746& 0.638& 1& N/A
\\
                                  & Gemma                                    & 0.14& N/A& 0.254& N/A& N/A& N/A
\\
                                  & Qwen                                    & 0.315& 0.586& 0.389& 0.386& N/A& N/A
\\ \hline
\multirow{4}{*}{AR-LSAT} & LLaMA                                   & 0.5& 0.432& 0.298& 0.33& 0.228& 0.281
\\
                                  & Mistral                                  & 0.585& 0.189& 0.289& 0.195& 0& 0
\\
                                  & Gemma                                    & 0.143& 0.158& 0.146& 0.131& N/A& N/A
\\
                                  & Qwen                                    & 0.22& 0.098& 0.079& 0.095& 0.062& 0.111\\     
\bottomrule
\end{tabular}
% }}
\label{tab:fn_to_tp}
\end{table*}

%% file: tables/tn_to_fp.tex
\begin{table*}[!t]
\centering
\caption{Response shift ratio by NASA on true negative samples (TN $\rightarrow$ FP ratio) in the rephrased datasets. High denotes high-confident and Low denotes low-confident.}
% {\resizebox{\textwidth}{!}{
\begin{tabular}{l|c||cccccc}
\toprule
\multirow{2}{*}{\textbf{Dataset}} & \multirow{2}{*}{\textbf{Model}} & \multicolumn{2}{c}{\textbf{Det-T}}                                   & \multicolumn{2}{c}{\textbf{Det-F}}                                   & \multicolumn{2}{c}{\textbf{Non-Det}}                                 \\
                                  &                               & \textbf{High} & \textbf{Low} & \textbf{High} & \textbf{Low} & \textbf{High} & \textbf{Low} \\ \hline
\multirow{4}{*}{MuSiQue} & LLaMA                                   & 0.057& 0.06& 0.144& 0.112& 0& 0.06
\\
                                  & Mistral                                  & 0.073& 0.075& 0.097& 0.08& 0.119& 0.032
\\
                                  & Gemma                                    & 0& N/A& 0.007& N/A& 0& N/A
\\
                                  & Qwen                                    & 0.009& 0.018& 0.015& 0.026& 0& N/A
\\ \hline
\multirow{4}{*}{GSM8K}   & LLaMA                                  & 0.587& 0.632& 0.516& 0.606& N/A& N/A
\\
                                  & Mistral                                  & 0.784& 0.65& 0.736& 0.652& N/A& N/A
\\
                                  & Gemma                                    & 0.271& N/A& 0.188& N/A& N/A& N/A
\\
                                  & Qwen                                    & 0.213& 0.208& 0.234& 0.303& N/A& N/A
\\ \hline
\multirow{4}{*}{AR-LSAT} & LLaMA                                   & 0.229& 0.349& 0.306& 0.247& 0.176& 0.21
\\
                                  & Mistral                                  & 0.208& 0.204& 0.248& 0.162& 0& 0.174
\\
                                  & Gemma                                    & 0.133& 0.083& 0.152& 0.134& N/A& N/A
\\
                                  & Qwen                                    & 0.109& 0.041& 0.13& 0.095& 0& 0\\     
\bottomrule
\end{tabular}
% }}
\label{tab:tn_to_fp}
\end{table*}

%% file: tables/q_only_results.tex
\begin{table*}[!t]
\centering
\caption{Ablation results of updated parameters (NASA / query projection only).}
{\resizebox{\textwidth}{!}{
\renewcommand{\arraystretch}{1.1}
\footnotesize
\begin{tabular}{@{}l|c||cccccc@{}}
\toprule
\textbf{Dataset} & \textbf{Model}                        & \textbf{Accuracy}   & \textbf{Precision} & \textbf{Recall}    & \textbf{F1}     & \textbf{NAS} & \textbf{ECE} \\ 
\hline
\multirow{3}{*}{StrategyQA} & LLaMA           & 0.877 / \textbf{0.877}        & 0.864 / \textbf{0.895}     &  \textbf{0.875} / 0.836     & \textbf{0.869} / 0.864  &  \textbf{55.0} / 96.2 & \textbf{0.102} / 0.128 \\
&Mistral       & 0.830 / \textbf{0.849}     & 0.802 / \textbf{0.846}   & \textbf{0.861} / 0.837     &  0.830 / \textbf{0.841}  & \textbf{34.3} / 71.9 & 0.116 / \textbf{0.114} \\
&Gemma                & 0.771 / \textbf{0.774}     & 0.907 / \textbf{0.907}    & 0.690 / \textbf{0.692}   &  0.784 / \textbf{0.785} & \textbf{71.8} / 79.2 & 0.162 / \textbf{0.161} \\
&Qwen                & \textbf{0.844} / 0.836     & 0.907 / \textbf{0.915}   & \textbf{0.742} / 0.716   &  \textbf{0.816} / 0.804 & \textbf{107.2} / 194.9 & \textbf{0.106} / 0.115 \\
\hline
\multirow{3}{*}{\parbox{1.7cm}{MuSiQue\\(Rephrased)}}   &LLaMA       & \textbf{0.757} / 0.741     & 0.815 / \textbf{0.844}    & \textbf{0.683} / 0.609    & \textbf{0.743} / 0.707 & \textbf{64.0} / 118.6 & \textbf{0.039} / 0.055 \\
& Mistral       & \textbf{0.707} / 0.704     & 0.802 / \textbf{0.821}    & \textbf{0.594} / 0.558    &  \textbf{0.682} / 0.664 & \textbf{110.7} / 185.9 & \textbf{0.212} / 0.217  \\
& Gemma                & 0.503 / \textbf{0.510}    & 0.834 / \textbf{0.836}    & \textbf{0.555} / 0.552    & \textbf{0.667} / 0.665 & \textbf{96.2} / 107.1 & 0.276 / \textbf{0.274} \\
&Qwen                & \textbf{0.639} / 0.622     & 0.858 / \textbf{0.873}    & \textbf{0.552} / 0.500   &  \textbf{0.672} / 0.636 & \textbf{282.9} / 405.7 & \textbf{0.200} / 0.215 \\
\hline
\multirow{3}{*}{\parbox{1.7cm}{GSM8K\\(Rephrased)}}& LLaMA            & 0.545 / \textbf{0.558}     & 0.531 / \textbf{0.551}    & \textbf{0.838} / 0.688    &  \textbf{0.650} / 0.612  & \textbf{74.7} / 123.6 & 0.170 / \textbf{0.105} \\
&Mistral       & \textbf{0.512} / 0.511     & 0.509 / \textbf{0.515}    & \textbf{0.782} / 0.463    &  \textbf{0.617} / 0.487 & \textbf{77.0} / 127.4 & \textbf{0.269} / 0.294 \\
&Gemma                & \textbf{0.524} / 0.517     & 0.541 / \textbf{0.542}   & 0.887 / \textbf{0.888}    &  0.672 / \textbf{0.673} & \textbf{66.2} / 74.7 & \textbf{0.346} / 0.349 \\
&Qwen                & \textbf{0.576} / 0.560     & 0.614 / \textbf{0.634}   & \textbf{0.738} / 0.606   &  \textbf{0.670} / 0.620 & \textbf{82.9} / 141.6 & 0.115 / \textbf{0.085} \\
\hline
\multirow{3}{*}{\parbox{1.7cm}{MATH\\(Rephrased)}}   &LLaMA            & 0.549 / \textbf{0.572}     & 0.534 / \textbf{0.562}    & \textbf{0.874} / 0.740     &  \textbf{0.663} / 0.639  & \textbf{60.8} / 103.6 & 0.175 / \textbf{0.118} \\
& Mistral       & 0.526 / \textbf{0.533}     & 0.522 / \textbf{0.549}     & \textbf{0.685} / 0.384    &  \textbf{0.592} / 0.452  & \textbf{109.6} / 152.8 & \textbf{0.260} / 0.299 \\
&Gemma               & \textbf{0.543} / 0.539     & 0.553 / \textbf{0.555}    & \textbf{0.871} / 0.867    &  \textbf{0.677} / 0.677 & \textbf{64.0} / 73.0 & 0.340 / \textbf{0.338} \\
&Qwen                & \textbf{0.581} / 0.549     & 0.617 / \textbf{0.639}    & \textbf{0.715} / 0.611   &  \textbf{0.662} / 0.624 & \textbf{103.1} / 187.5 & \textbf{0.162} / 0.168 \\
\hline
\multirow{3}{*}{\parbox{1.7cm}{AR-LSAT\\(Rephrased)}}   & LLaMA            & \textbf{0.518} / 0.510     & \textbf{0.537} / 0.536   & \textbf{0.444} / 0.337   &  \textbf{0.486} / 0.414 & \textbf{63.6} / 109.2 & \textbf{0.227} / 0.245 \\
& Mistral      & \textbf{0.530} / 0.526     &  0.553 / \textbf{0.556}     & \textbf{0.447} / 0.382    &  \textbf{0.494} / 0.453  & \textbf{80.02} / 121.3 & \textbf{0.355} / 0.382 \\
& Gemma                & 0.510 / \textbf{0.514}     & 0.523 / \textbf{0.526}    & 0.534 / \textbf{0.551}    &  0.528 / \textbf{0.539} & \textbf{68.0} / 75.6 & \textbf{0.372} / 0.383 \\
&Qwen                & 0.496 / \textbf{0.559}     & 0.528 / \textbf{0.643}    & 0.182 / \textbf{0.574}   &  0.270 / \textbf{0.607} & 207.3 / \textbf{204.4} & \textbf{0.359} / 0.384 \\
\bottomrule
\end{tabular}
}}
\label{tab:appendix:q_only_results}
\end{table*}

%% file: tables/random_head_tuning.tex
\begin{table*}[!t]
\centering
\caption{Ablation results of attention head selection (NASA / random attention heads).}
{\resizebox{\textwidth}{!}{
\renewcommand{\arraystretch}{1.1}
\footnotesize
\begin{tabular}{@{}l|c||cccccc@{}}
\toprule
\textbf{Model} & \textbf{Dataset}                        & \textbf{Accuracy}   & \textbf{Precision} & \textbf{Recall}    & \textbf{F1}     & \textbf{NAS} & \textbf{ECE} \\ 
\hline
\multirow{4}{*}{LLaMA} & StrategyQA           &  \textbf{0.877} / 0.870        & 0.864 / \textbf{0.924}     &  \textbf{0.875} / 0.789     & \textbf{0.869} / 0.851  &  55.0 / $\mathbf{-507.6}$ & \textbf{0.102} / 0.176 \\
&MuSiQue       & \textbf{0.757} / 0.722     & 0.815 / \textbf{0.862}   & \textbf{0.683} / 0.547     &  \textbf{0.743} / 0.669  & 64.0 / $\mathbf{-956.7}$ & \textbf{0.039} / 0.076 \\
&GSM8K                & 0.545 / \textbf{0.554}     & 0.531 / \textbf{0.570}    & \textbf{0.838} / 0.482   &  \textbf{0.650} / 0.522 & 74.7 / $\mathbf{-433.8}$  & 0.170 / \textbf{0.109} \\
&MATH                & 0.549 / \textbf{0.562}     & 0.534 / \textbf{0.573}   & \textbf{0.874} / 0.561   &  \textbf{0.663} / 0.567 & 60.8 / $\mathbf{-390.1}$   & 0.175 / \textbf{0.095} \\
&AR-LSAT                & 0.518 / 0.518     & 0.537 / \textbf{0.570}   & \textbf{0.444} / 0.249   &  \textbf{0.486} / 0.346 & 63.6 / $\mathbf{-817.5}$ & \textbf{0.227} / 0.259 \\
\hline
\multirow{4}{*}{Mistral}   &StrategyQA       & 0.830 / \textbf{0.849}     & 0.802 / \textbf{0.888}    & \textbf{0.861} / 0.784    & 0.830 / \textbf{0.833} & \textbf{34.3} / 750.8 & 0.116 / \textbf{0.112} \\
& MuSiQue       & \textbf{0.707} / 0.679     & 0.802 / \textbf{0.850}    & \textbf{0.594} / 0.483    &  \textbf{0.682} / 0.616 & \textbf{110.7} / 1405.1  & \textbf{0.212} / 0.244 \\
& GSM8K                & 0.512 / \textbf{0.518}    & 0.509 / \textbf{0.539}    & \textbf{0.782} / 0.287    & \textbf{0.617} / 0.374 & \textbf{77.0} / 1055.0  & \textbf{0.269} / 0.339 \\
&MATH                & 0.525 / \textbf{0.528}     & 0.522 / \textbf{0.563}    & \textbf{0.685} / 0.266   &  \textbf{0.592} / 0.361 & \textbf{109.6} / 1042.2 & \textbf{0.260} / 0.354 \\
&AR-LSAT                & \textbf{0.530} / 0.514     & \textbf{0.553} / 0.551    & \textbf{0.447} / 0.290   &  \textbf{0.494} / 0.380 & \textbf{80.0} / 1015.0 & \textbf{0.355} / 0.396 \\
\hline
\multirow{4}{*}{Gemma}   &StrategyQA       & 0.771 / \textbf{0.780}     & 0.907 / \textbf{0.922}    & \textbf{0.690} / 0.678    & \textbf{0.784} / 0.782 & \textbf{71.8} / 104.7 & \textbf{0.162} / 0.165 \\
& MuSiQue       & 0.503 / \textbf{0.532}     & \textbf{0.834} / 0.829    & \textbf{0.555} / 0.534    &  \textbf{0.667} / 0.650 & \textbf{96.2} / 123.8  & \textbf{0.276} / 0.279 \\
& GSM8K                & 0.524 / \textbf{0.533}    & \textbf{0.541} / 0.540    & \textbf{0.887} / 0.855    & \textbf{0.672} / 0.662 & \textbf{66.2} / 103.5  & \textbf{0.346} / 0.357 \\
&MATH                & 0.543 / \textbf{0.547}     & 0.553 / \textbf{0.554}    & \textbf{0.871} / 0.835   &  \textbf{0.677} / 0.666 & \textbf{64.0} / 105.4 & 0.340 / \textbf{0.331} \\
&AR-LSAT                & 0.510 / \textbf{0.513}     & 0.523 / \textbf{0.529}    & \textbf{0.534} / 0.489   &  \textbf{0.528} / 0.508 & \textbf{68.0} / 98.6 & \textbf{0.372} / 0.374 \\
\hline
\multirow{4}{*}{Qwen}   &StrategyQA       & \textbf{0.844} / 0.832     & 0.907 / \textbf{0.917}    & \textbf{0.742} / 0.705    & \textbf{0.816} / 0.797 & \textbf{107.2} / 202.0 & \textbf{0.106} / 0.116 \\
& MuSiQue       & \textbf{0.639} / 0.627     & 0.858 / \textbf{0.880}    & \textbf{0.552} / 0.505    &  \textbf{0.672} / 0.642 & \textbf{282.9} / 392.8  & \textbf{0.200} / 0.221 \\
& GSM8K                & \textbf{0.576} / 0.540    & 0.614 / \textbf{0.637}    & \textbf{0.738} / 0.562    & \textbf{0.670} / 0.597 & \textbf{82.9} / 195.7  & 0.115 / \textbf{0.091} \\
&MATH                & \textbf{0.581} / 0.563     & 0.617 / \textbf{0.652}    & \textbf{0.715} / 0.556   &  \textbf{0.662} / 0.600 & \textbf{103.1} / 201.3 & \textbf{0.162} / 0.168 \\
&AR-LSAT                & \textbf{0.496} / 0.493     & 0.528 / \textbf{0.530}    & \textbf{0.182} / 0.117   &  \textbf{0.270} / 0.192 & \textbf{207.3} / 299.4 & \textbf{0.359} / 0.392 \\
\bottomrule
\end{tabular}
}}
\label{tab:appendix:random_head_results}
\end{table*}

%% file: figures/conf_hists_others_.tex
\begin{figure*}[!t]
  \includegraphics[width=\textwidth]{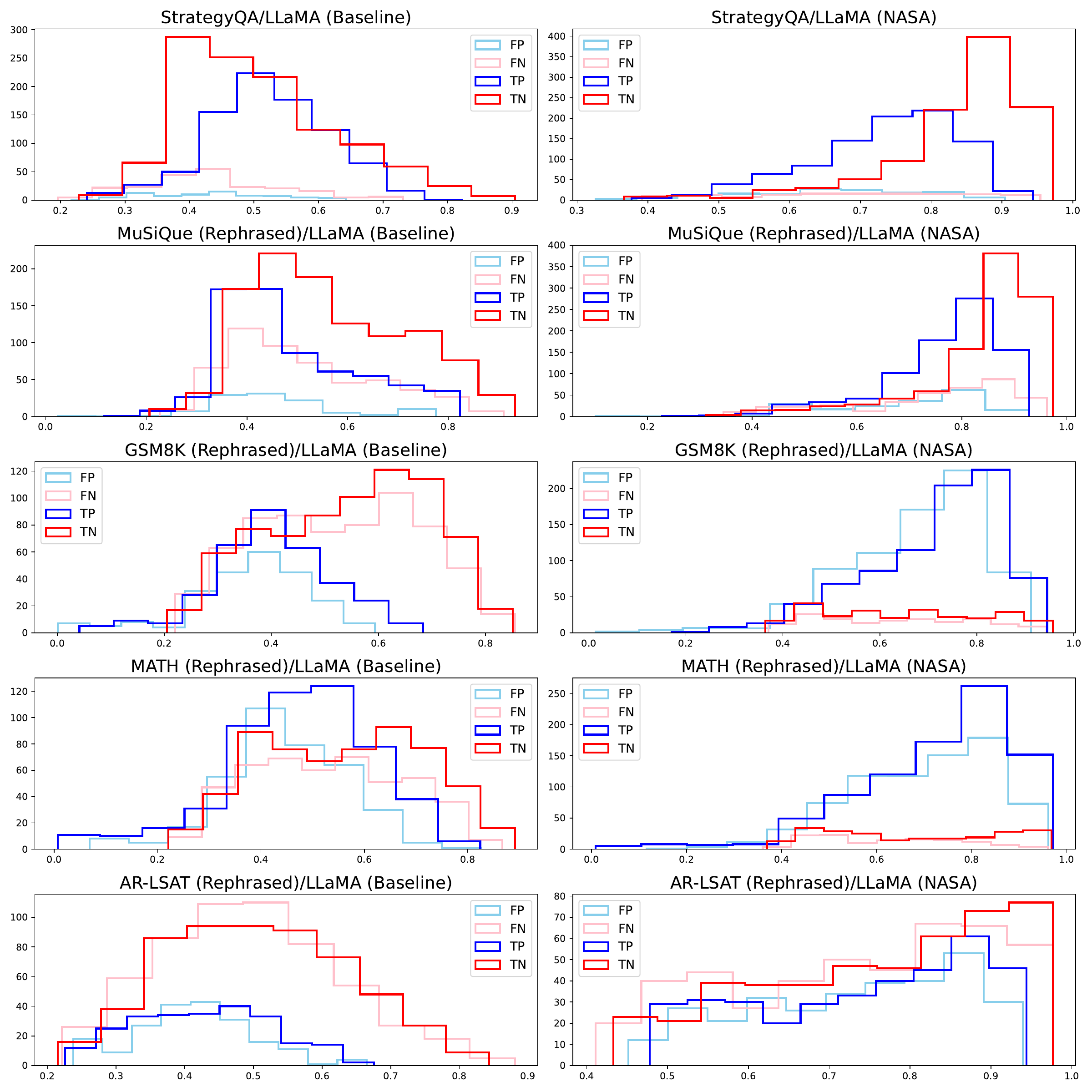}
  \caption{Confidence histograms of LLaMA before and after NASA.}
  \label{fig:conf_hists_llama}
\end{figure*}

%% file: figures/attn_heads_llama_.tex
\begin{figure*}[!t]
  \includegraphics[width=\textwidth]{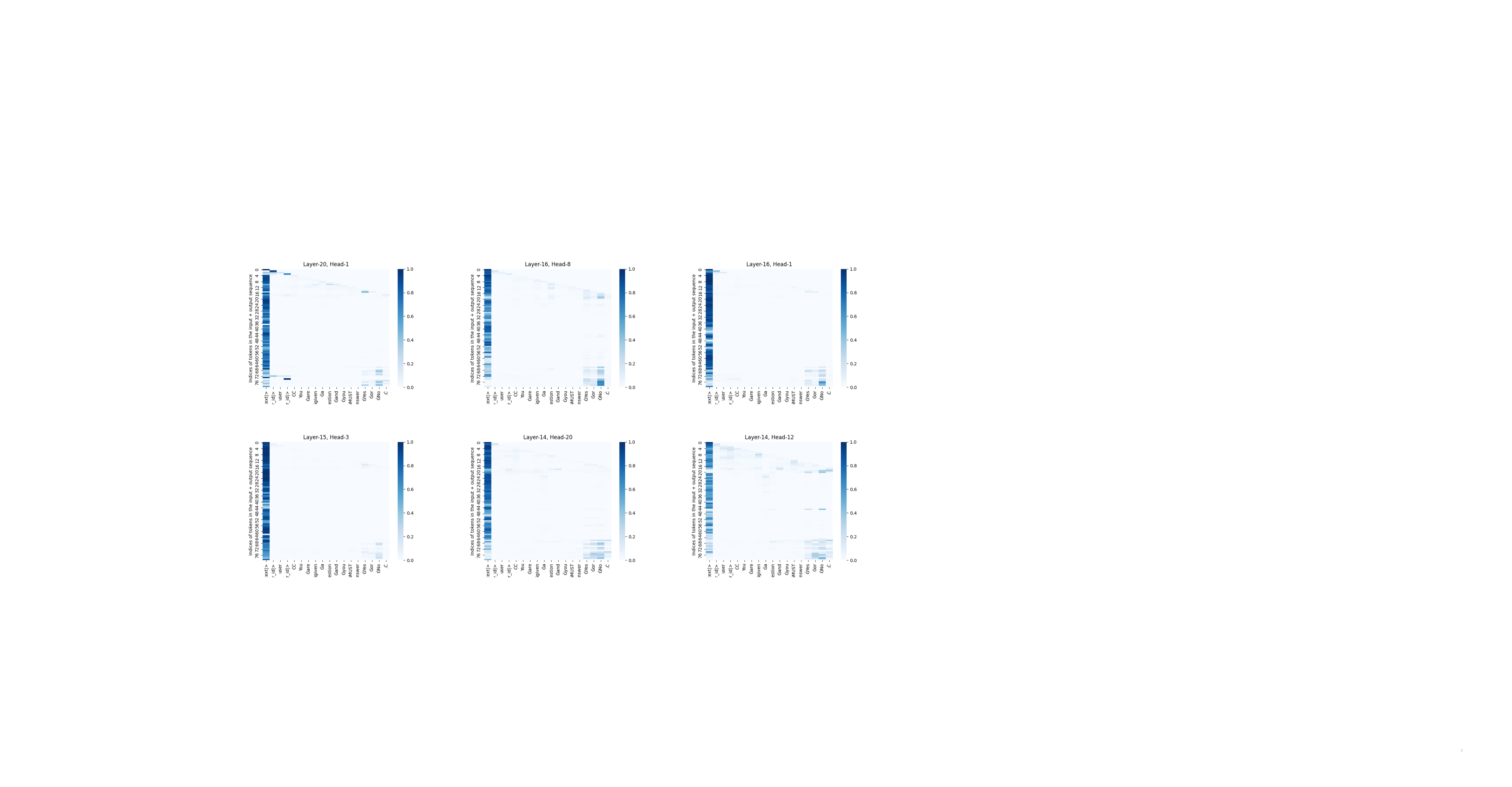}
  \caption{Examples of attention heads of LLaMA for a GSM8K dataset sample.}
  \label{fig:attn_head_llama}
\end{figure*}

%% file: figures/attn_heads_nasa_.tex
\begin{figure*}[!t]
  \includegraphics[width=\textwidth]{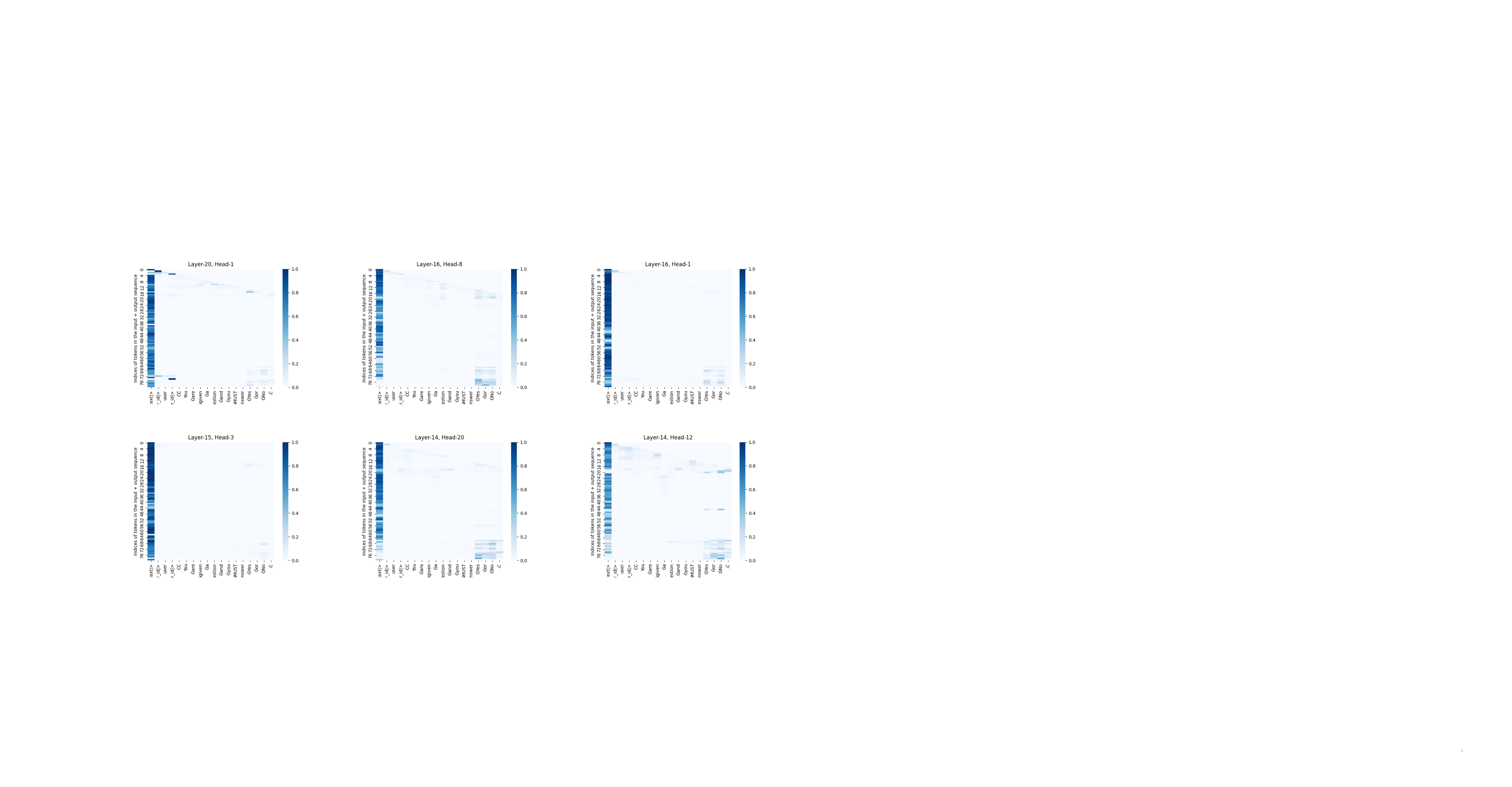}
  \caption{Examples of attention heads of LLaMA model with NASA for a GSM8K dataset sample.}
  \label{fig:attn_head_nasa}
\end{figure*}